\documentclass[10pt,journal,compsac]{IEEEtran}
\usepackage[table]{xcolor}
\usepackage{epsfig}
\usepackage{graphicx}
\usepackage{amsmath}
\usepackage{amssymb}
\usepackage{enumitem}

\usepackage{multirow}
\usepackage{tabularx}
\usepackage{tcolorbox}
\usepackage{dsfont}
\usepackage{url}
\usepackage[tight]{subfigure}
\usepackage{color}
\usepackage{subfigure}
\usepackage{amsthm}
\usepackage[export]{adjustbox}
\usepackage{comment}
\usepackage{diagbox}

\usepackage[utf8]{inputenc} 
\usepackage[T1]{fontenc}    
\usepackage{url}            
\usepackage{booktabs}       
\usepackage{amsfonts}       
\usepackage{nicefrac}       
\usepackage{microtype}      
\usepackage{pifont}
\usepackage{balance}
\usepackage[rightcaption]{sidecap}
\usepackage[linesnumbered,titlenumbered,ruled,vlined,commentsnumbered,noend]{algorithm2e}

\def\argmax{\operatornamewithlimits{arg\,max}}

\usepackage{xspace}
\makeatletter
\DeclareRobustCommand\onedot{\futurelet\@let@token\@onedot}
\def\@onedot{\ifx\@let@token.\else.\null\fi\xspace}
\def\eg{\emph{e.g}\onedot} 
\def\ie{\emph{i.e}\onedot} 
 
\def\etc{\emph{etc}\onedot}

\makeatother

\definecolor{royalblue}{RGB}{65,105,225} %
\definecolor{lightblue}{RGB}{170,224,250} %
\definecolor{lightgreen}{RGB}{196,223,155}
\definecolor{lightyellow}{RGB}{254,247,153}
\usepackage[pagebackref=true,breaklinks=true,colorlinks,citecolor=royalblue,bookmarks=false]{hyperref}

\usepackage{scalerel}
\usepackage{tikz}
\usetikzlibrary{svg.path}
\definecolor{orcidlogocol}{HTML}{A6CE39}
\tikzset{
  orcidlogo/.pic={
    \fill[orcidlogocol] svg{M256,128c0,70.7-57.3,128-128,128C57.3,256,0,198.7,0,128C0,57.3,57.3,0,128,0C198.7,0,256,57.3,256,128z};
    \fill[white] svg{M86.3,186.2H70.9V79.1h15.4v48.4V186.2z}
                 svg{M108.9,79.1h41.6c39.6,0,57,28.3,57,53.6c0,27.5-21.5,53.6-56.8,53.6h-41.8V79.1z M124.3,172.4h24.5c34.9,0,42.9-26.5,42.9-39.7c0-21.5-13.7-39.7-43.7-39.7h-23.7V172.4z}
                 svg{M88.7,56.8c0,5.5-4.5,10.1-10.1,10.1c-5.6,0-10.1-4.6-10.1-10.1c0-5.6,4.5-10.1,10.1-10.1C84.2,46.7,88.7,51.3,88.7,56.8z};
  }
}
\newcommand\orcidicon[1]{\href{https://orcid.org/#1}{\mbox{\scalerel*{
\begin{tikzpicture}[yscale=-1,transform shape]
\pic{orcidlogo};
\end{tikzpicture}
}{|}}}}

\begin{document}

\title{Dodging DeepFake Detection via Implicit Spatial-Domain Notch Filtering} 

\author{Yihao~Huang\,\orcidicon{0000-0002-5784-770X},
        Felix~Juefei-Xu\,\orcidicon{0000-0002-0857-8611},~\IEEEmembership{Member,~IEEE,}
        Qing~Guo$^\dagger$\,\orcidicon{0000-0003-0974-9299},~\IEEEmembership{Member,~IEEE,}
        Yang~Liu\,\orcidicon{0000-0001-7300-9215},~\IEEEmembership{Senior~Member,~IEEE,}
        and~Geguang~Pu\,\orcidicon{0000-0001-9750-8334}
\thanks{Geguang Pu is supported by National Key Research and Development Program (2020AAA0107800), and Shanghai Collaborative Innovation Center of  Trusted Industry Internet Software. This research is supported by the National Research Foundation, Singapore, and DSO National Laboratories under the AI Singapore Programme (AISG Award No: AISG2-GC-2023-008), the A*STAR Centre for Frontier AI Research, the National Research Foundation, Singapore, and the Cyber Security Agency under its National Cybersecurity R\&D Programme (NCRP25-P04-TAICeN), and NRF Investigatorship NRF-NRFI06-2020-0001. Any opinions, findings and conclusions or recommendations expressed in this material are those of the author(s) and do not reflect the views of National Research Foundation, Singapore and Cyber Security Agency of Singapore.

Yihao~Huang and Yang~Liu are with Nanyang Technological University, Singapore. Felix~Juefei-Xu is with New York University, USA. Qing~Guo is with the Institute of High Performance Computing (IHPC) and Centre for Frontier AI Research (CFAR), Agency for Science, Technology and Research (A*STAR), Singapore. Geguang~Pu is with 1) East China Normal University and 2) Shanghai Industrial Control Safety Innovation Technology Co., LTD, China. $\dagger$ Qing Guo is the corresponding author (tsingqguo@ieee.org).}
}


\maketitle
\begin{abstract}
The current high-fidelity generation and high-precision detection of DeepFake images are at an arms race. We believe that producing DeepFakes that are highly realistic and ``detection evasive'' can serve the ultimate goal of improving future generation DeepFake detection capabilities. In this paper, we propose a simple yet powerful pipeline to reduce the artifact patterns of fake images without hurting image quality by performing implicit spatial-domain notch filtering. We first demonstrate that frequency-domain notch filtering, although famously shown to be effective in removing periodic noise in the spatial domain, is infeasible for our task at hand due to the manual designs required for the notch filters. We, therefore, resort to a learning-based approach to reproduce the notch filtering effects, but solely in the spatial domain. We adopt a combination of adding overwhelming spatial noise for breaking the periodic noise pattern and deep image filtering to reconstruct the noise-free fake images, and we name our method \textbf{DeepNotch}. Deep image filtering provides a specialized filter for each pixel in the noisy image, producing filtered images with high fidelity compared to their DeepFake counterparts. Moreover, we also use the semantic information of the image to generate an adversarial guidance map to add noise intelligently. 
Our large-scale evaluation on 3 representative DeepFake detection methods (tested on 16 types of DeepFakes) has demonstrated that our technique significantly reduces the accuracy of these 3 fake image detection methods, 36.79\% on average and up to 97.02\% in the best case.

\end{abstract}
\begin{IEEEkeywords}
DeepFake, DeepFake Evasion, DeepFake Detection
\end{IEEEkeywords}

\IEEEpeerreviewmaketitle

\section{Societal Impact}
Our proposed method, just like many other high-fidelity image synthesis or DeepFake generation methods, if maliciously used by an adversary, may cause harm to the integrity of digital media and fuel the dissemination of misinformation and disinformation. The study herewithin attempts to expose potential vulnerabilities of the deployed defense mechanism with the goal of ultimately improving it by presenting a stronger contender. Our method aims at improving the fidelity of DeepFake images and, more importantly, exposing the problems of existing DeepFake detection methods, and we hope that the found vulnerabilities can help improve future generation DeepFake detection.
\begin{figure}
	\centering 
	\includegraphics[width=\columnwidth]{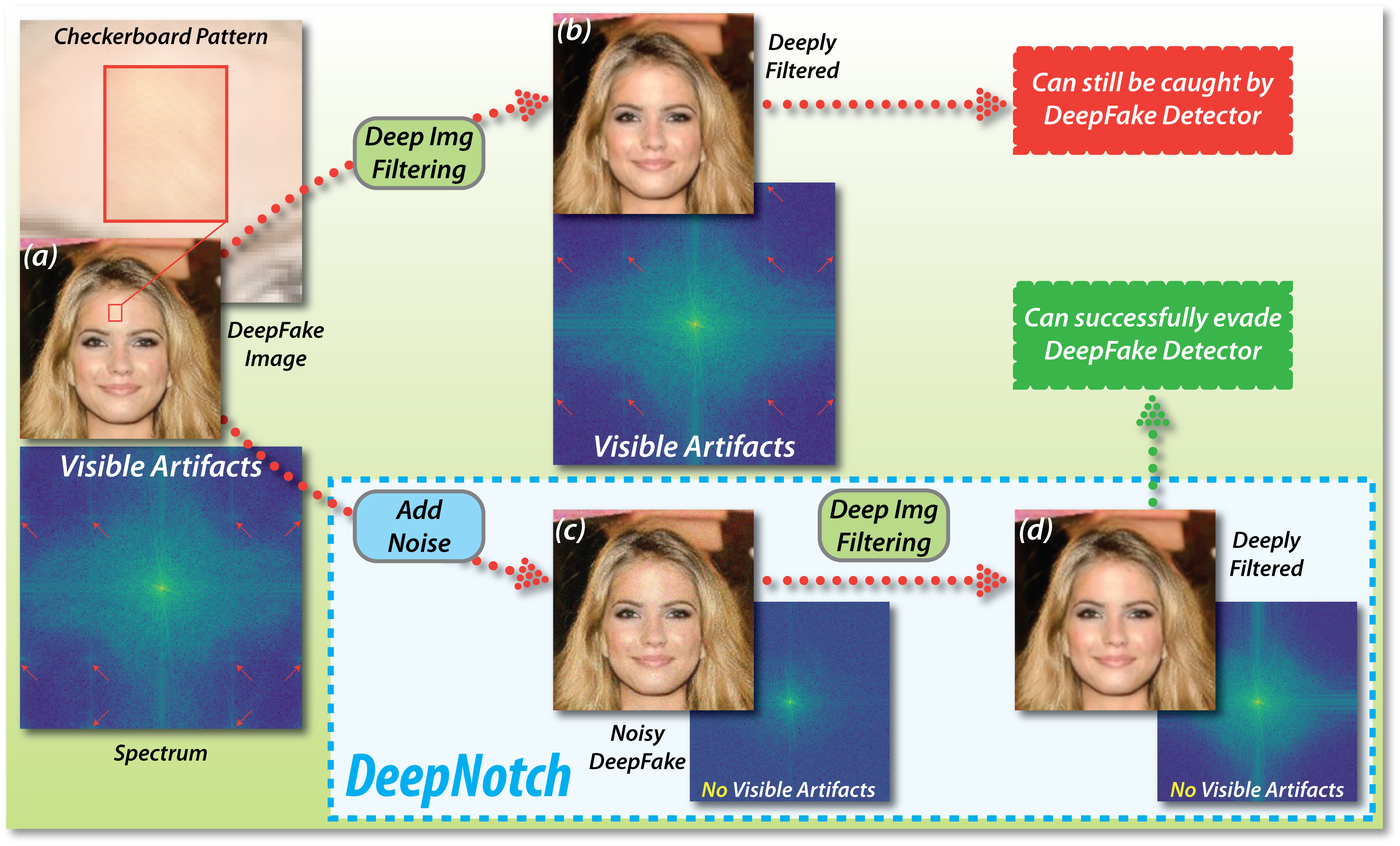}
	\caption{(a) DeepFake shows the checkerboard pattern and visible spike artifact in its spectrum (red arrows). (Top) Using deep image filtering to directly retouch the DeepFake image will still leave artifacts and it can be caught by DeepFake detectors (b). (Bottom) Performing implicit notch filtering with \textbf{DeepNotch}. By adding noise to the DeepFake, we can successfully reduce the artifact pattern (c). Then, deep image filtering is able to restore the image quality without adding artifacts (d), thus successfully evading DeepFake detectors.}
	\label{fig:teaser}
\end{figure}

\section{Introduction}\label{sec:intro}

Fake images produced by the generative adversarial network (GAN) and its variants can now render both photo-realistic and high-fidelity effects, \emph{a.k.a.} DeepFakes. However, the state-of-the-art (SOTA) GAN-based fake image generation methods are still imperfect, which stems from the upsampling modules in their decoders. 
In particular, existing upsampling methods of GANs, \eg, transpose convolution, unpooling, and interpolation, often inevitably introduce artifact patterns to the generated fake images, occurring in both spatial-domain and frequency-domain (\eg, Fourier spectrum) representations. For example, checkerboard patterns are typical textures that may be left in the generated fake images. Similarly, typical artifact patterns in the spectrum of fake images are also discussed in \cite{zhang2019detecting,frank2020leveraging}.
Through leveraging the potential artifact patterns as the clue, quite a few DeepFake detection methods have been proposed \cite{yu2019attributing,wang2020cnn,zhang2019detecting,frank2020leveraging,ni2022core,dong2022think,jeong2022frepgan,li2021frequency,liu2021spatial,zhao2021multi,dzanic2020fourier,qian2020thinking}. These methods largely fall into three categories according to their inputs: \emph{image-based}, \emph{fingerprint-based}, and \emph{spectrum-based}, most of which have demonstrated their effectiveness in successfully detecting SOTA DeepFakes.

To generate more photo-realistic and high-fidelity fake images, one promising direction is to reduce the artifact patterns introduced in fake images.
Along this line, in this paper, we propose a pipeline \textbf{DeepNotch} to perform implicit spatial-domain notch filtering on the DeepFake images to make them more \emph{realistic} and detection evasive (Fig.~\ref{fig:teaser}).

Our key observation and intuition are that adding noise into fake images can reduce and, to some extent, destroy the artifact patterns in both spatial and frequency domains of the fake images. This observation paves a new path to design a novel deep image filtering method to retouch the fake images, reducing these fake artifacts.
In particular, the deep image filtering method enables us to effectively produce a specified kernel for each pixel of an image in a unique way. More importantly, such a procedure does not introduce unwanted extra artifacts.
Through jointly combining the operations of adding noise and deep image filtering, {DeepNotch} can not only reduce the artifact patterns but also maintain the quality of fake images.
Furthermore, to make the noise even ``smarter'', we also propose an adversarial guided map to pinpoint the tentative locations to add the noises. To our best knowledge, this is the first work that creates high-quality fake images from the retouch and refinement perspective based on deep learning (DL), which also exhibits strong capability in avoiding state-of-the-art DeepFake detectors. Our new findings of this paper could be potentially helpful and guide us to think about how to design more effective DeepFake detection methods.

In summary, the contributions as listed below:
\begin{itemize}[leftmargin=*]
    \item 
    We propose the first DL-based fake image retouch method to reduce the artifact by performing implicit notch filtering. The reconstructed fake images are both photo-realistic and have a strong capability in bypassing SOTA fake detectors. To further improve the effectiveness of DeepNotch, we propose a novel semantic-aware localization method to pinpoint the places for noise addition.
    \item 
    We perform a large-scale evaluation on 3 SOTA fake image detection methods and the fake images are generated by a total of 16 popular DeepFake generation methods. In particular, our reconstructed fake images can significantly reduce the fake detection accuracy of DeepFake detectors and they exhibit a high level of fidelity compared to their original fake image counterparts.
    \item 
    Our method indicates that existing detection methods highly leverage the information of artifact patterns for fake detection. The observation also raises an open question of how to propose more general DeepFake detection methods.
\end{itemize}

\section{Related Work}\label{sec:related}

\subsubsection{GAN-based image generation \& manipulation}
Since its advent, GANs \cite{goodfellow2014generative} have been extensively studied with many GAN-based image generation methods proposed \cite{radford2015unsupervised,arjovsky2017wasserstein,gulrajani2017improved,perarnau2016invertible,karras2017progressive,karras2019style,karras2020analyzing,miyato2018spectral,li2017mmd,he2019attgan,choi2018stargan,liu2019stgan,choi2020stargan,gao2021high}. Specifically, IcGAN first encodes real images into the latent space and then changes the latent codes corresponding to different facial properties. After that, it decodes the changed latent codes to fake face images. In recent years, there have been some GAN-based image generation and manipulation methods that put emphasis on stably generating images with high-resolution and controllable face attributes.
ProGAN \cite{karras2017progressive} proposes to synthesize high-resolution images through growing both the generator and discriminator, which achieves efficient training performance. StyleGAN \cite{karras2019style} designs a different generator architecture that leads to an automatically learned separation of high-level attributes and stochastic variation in the generated images. 
Based on StyleGAN, StyleGAN2 \cite{karras2020analyzing} further proposes improvements on both model architecture and training methods for higher image generation quality. 
SNGAN \cite{miyato2018spectral} proposes a lightweight normalization method to stabilize and enhance the training of the discriminator.  MMDGAN \cite{li2017mmd} combines the key ideas in both the generative moment matching network (GMMN) and GAN. 
\begin{figure}
\centering 
	\includegraphics[width=\columnwidth]{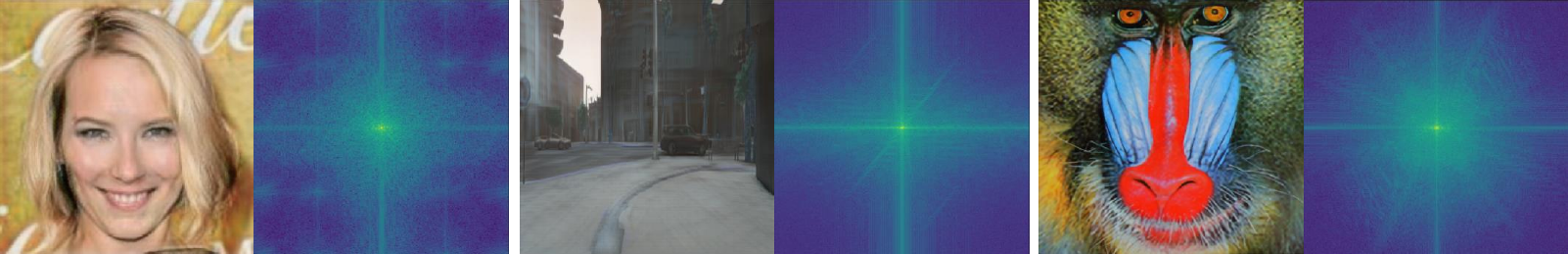}
	\caption{(L-R) Fake image produced by StarGAN \cite{choi2018stargan}, CRN \cite{shi2016end}, and SAN \cite{dai2019second} and their corresponding spectrum.}
	\label{fig:spectrum_of_different_GAN}
\end{figure}
Some GANs are also specially designed to manipulate the facial attributes of real images.
AttGAN \cite{he2019attgan} adopts an attribute detection constraint to the generated image instead of latent representation to guarantee the correct change of desired attributes. STGAN \cite{liu2019stgan} selectively analyzes the difference between target and source attribute vectors and adaptively modifies the encoder feature for enhanced attribute editing. 
StarGAN \cite{choi2018stargan} proposes a novel approach that can perform image-to-image translations for multiple domains using only a single model. 
%
%

\subsubsection{DeepFake detection methods}\label{DeepFake Detection Methods}
\cite{tolosana2020deepfakes} and \cite{juefei2021countering} recently conduct comprehensive surveys on the DeepFake detection methods \cite{wang2019fakespotter,arxiv20_deeprhythm,wang2020cnn,afchar2018mesonet,nguyen2019capsule,nguyen2019multi,yu2019attributing,zhang2019detecting,frank2020leveraging,arxiv20_deepsonar,fernandes2019predicting,chen2018neural,qi2020deeprhythm}. Overall, they surveyed hundreds of papers that focus on DeepFake detection, most of which proposed CNN-based deep learning methods to solve the detection problem. The methods mainly leverage image clues or biological signal clues to address the detection tasks. Some work such as \cite{fraga2020fake} (which provide a comprehensive overview by leveraging distributed ledger technologies (DLT) to combat digital deception) and \cite{hasan2019combating} (leverage blockchain to trace and track the source of multimedia which provides insight for combating DeepFake videos) are not regular detection based on pure image analysis are not considered in our work.

The biological signal exhibits a clear signal for distinguishing between real and fake. The biological signals revealed in the real faces videos are natural and realistic while is low-quality in fake videos. Early works study the irregular eye blinking~\cite{li2018ictu}, the mismatch facial landmarks~\cite{yang2019exposing}, \etc. However, these visual inconsistencies could be easily removed in the advanced DeepFakes. In recent years, some works \cite{fernandes2019predicting,chen2018neural,qi2020deeprhythm} took heartbeats as the clue to classify the videos. Some work \cite{mittal2020emotions,chugh2020not,Agarwal_2020_CVPR_Workshops,hu2021detecting} detect DeepFake according to the inconsistency of visual and audio. Since the biological signal-based detection methods mainly focus on video and are hardly accessible on images, thus our image reconstruction method does not take them into consideration.

For detection methods that consider image clues, they largely fall into three categories depending on their feature inputs: \emph{image-based methods} \cite{wang2019fakespotter,wang2020cnn,afchar2018mesonet,nguyen2019capsule,nguyen2019multi}, \emph{fingerprint-based methods} \cite{marra2019gans,yu2019attributing}, and \emph{spectrum-based methods} \cite{zhang2019detecting,dzanic2020fourier}. 
As many CNN did, image-based methods perform fake detection directly on the original images (as inputs). 
Fingerprint-based methods follow the intuition that different GANs have various fingerprints. Through analyzing the features of GAN fingerprints, they can successfully detect fake images in many cases. 
Spectrum-based methods take another perspective, which leverages spectrum as the input of their network for more effective fake image detection, with the intuition that DeepFake artifacts are manifested as replications of spectra in the frequency domain.

\subsubsection{Diffusion Model}\label{Diffusion_models}
Recently, diffusion models (DM) \cite{ho2020denoising,sd,rombach2022high} are versatile tools with a wide array of applications, such as image generation \cite{song2021scorebased}, image translation \cite{meng2022sdedit}, voice synthesis \cite{liu2022diffsinger} and adversarial defense \cite{nie2022DiffPure}. Diffusion model performing iteratively adding Gaussian noise on an image as a forward process and denoising to restore the image by a backward process. From this point of view, our work shows similarity with the diffusion model in that we first add Gaussian noise to the fake image and then denoise with the deep learning-based method. Since DM-based image-to-image translation \cite{meng2022sdedit} shows the ability to transform the image from a specific domain to another domain, we think it somewhat verifies the reasonability of our work to transform the fake image from a fake domain to a real domain.

\section{Methodology}\label{sec:method}

\subsection{Motivation}\label{sec:motivation}

It is widely observed that various GAN-based image generation methods leave some footprints of artifacts in the image's Fourier spectrum, \eg, bright spots that are symmetric about the origin, star-shaped line segments symmetrically shooting from the origin, \etc, as shown in Fig.~\ref{fig:spectrum_of_different_GAN}. The artifacts correspond to \emph{fake patterns} in the spatial domain. For various DeepFake generation methods, such artifacts usually appear to be checkerboard-like in the spatial domain, producing several pairs of origin-symmetric bright spots (spikes) in the Fourier spectrum, indicating that the pattern contains more than just one sinusoidal component.

\begin{figure}
	\centering 
	\includegraphics[width=\columnwidth]{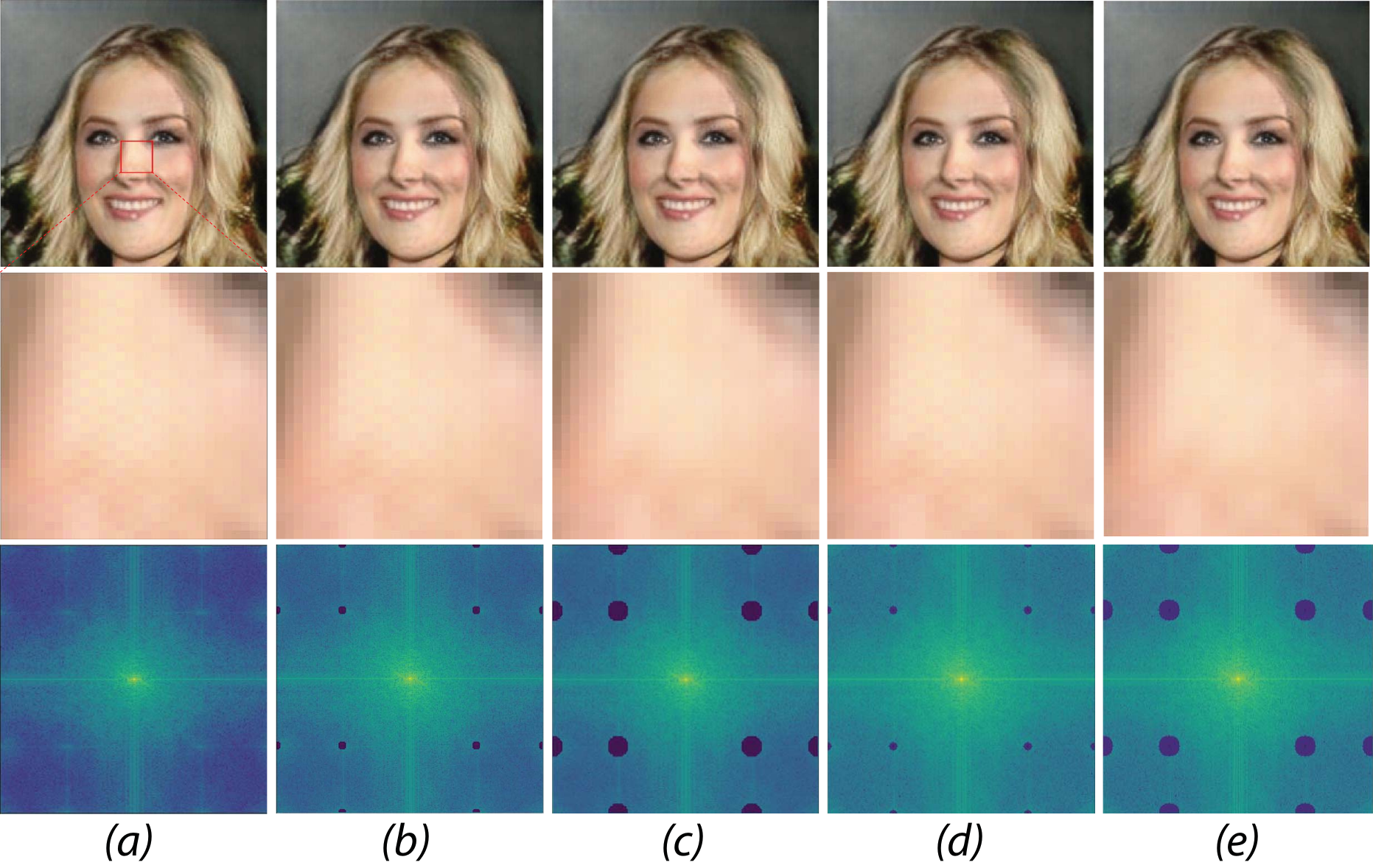}
	\caption{(a) StarGAN \cite{choi2018stargan} DeepFake. The zoomed-in area shows a clear checkerboard pattern and the spectrum also shows artifacts. (b) An ideal notch filter $(r=4)$ is applied to eliminate frequency-domain noise. The corresponding partial and full images exhibit fewer checkerboard patterns. (c) An ideal notch filter $(r=10)$ is applied. The checkerboard patterns almost disappear. (d-e) We replace the ideal notch filter with a Gaussian notch filter $(\sigma=1)$, leading to similar performances as (b-c).}
	\label{fig:notch_for_toy_example}
\end{figure}

In image processing literature \cite{gonzales2002digital}, a common and effective way of removing or mitigating fake patterns in the spatial domain is to apply frequency-domain filtering with a notch filter. A notch filter (or notch reject filter in this case) can be of various shapes, sizes, and orientations, and applied at various spectrum locations, \etc. Depending on how the artifacts look in the spectrum, the notch filter can be designed manually to maximally reduce the energy surrounding the artifacts. In our case, \eg, the exhibited artifacts are bright spots symmetrically positioned in the spectrum (see Fig.~\ref{fig:notch_for_toy_example} (a)). Therefore, any circular-shaped notch filter such as disk filters, Butterworth filters, or Gaussian filters, will suffice to encapsulate the artifact spikes. In Fig.~\ref{fig:notch_for_toy_example} (b-e), applying disk and Gaussian notch filters at specific spectrum locations can effectively remove the checkerboard pattern in the spatial domain. However, we will explain why notch filtering in the frequency domain is not a feasible solution for automatic detection-evasion.

\begin{figure}
	\centering 
	\includegraphics[width=\columnwidth]{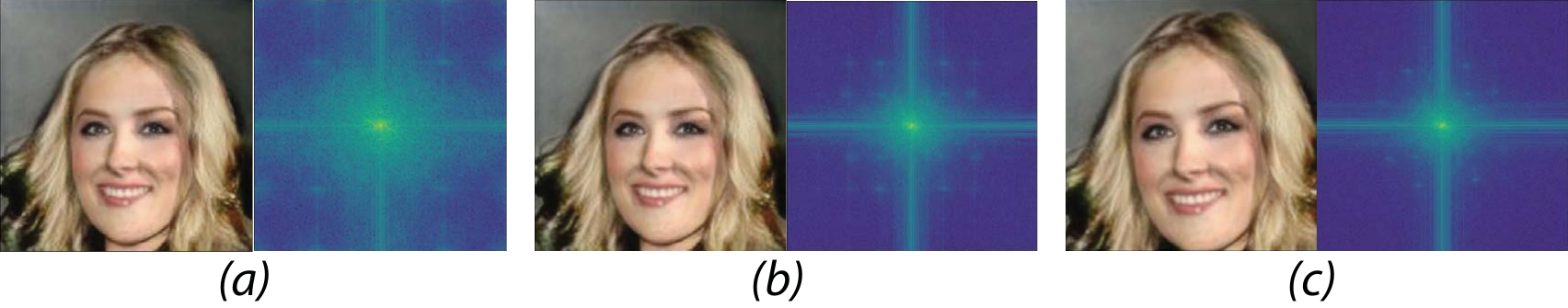}
	\caption{(a) StarGAN \cite{choi2018stargan} DeepFake of size 256 $\times$ 256 and its spectrum. The spectrum has clear artifact patterns. (b) We resize the image into size 512 $\times$ 512 and find that the artifact patterns of its spectrum shift obviously. (c) Based on the resize process in (b), we then rotate the image by 5 degrees and enlarge the image to 1.1 times. The artifact patterns in the spectrum also rotate.}
	\label{fig:spectrum_for_toy_example}
\end{figure}

Here are the main challenges hindering a successful application of notch filtering in the frequency domain. \ding{182} Different GANs may result in different spikes (bright spots) pairs in the spectrum and at different locations. Therefore, human-in-the-loop localization of the spikes is required, rendering it infeasible. \ding{183} Spikes are the simplest types of artifacts, and there are artifacts with many complicated patterns, \eg, requiring rectangular notch filters with notch openings positioned at particular locations, and the filter positioned at particular orientations. All these efforts require human involvement. \ding{184} Even if some forgery method produces quite consistent artifacts, \eg, with fixed spike location, the spikes in the image spectrum may rotate or shift due to simple geometric transformation of the generated image (see a toy example in Fig.~\ref{fig:spectrum_for_toy_example}), where slight rotation and scaling transformation are applied to the image and the spikes in the spectrum are relocated. Automatically designing notch filters to account for possible geometric perturbation is infeasible in general. \ding{185} For partially exhibited fake patterns in the spatial domain, the spikes are usually energy-spread into a cloud shape in the spectrum, making the determination of notch sizes less definite. 

For these reasons, it is usually not easy to automatically design a frequency-domain notch filter based on the fake patterns in the spatial domain. Also, there is usually not a corresponding convolution-based spatial-domain filter to achieve fake pattern removal due to the nature of the notch filter \cite{gonzales2002digital}. Therefore, notch filtering, although very effective in removing fake patterns in the spatial domain, without human involvement, it has limited usage in tackling the DeepFake detection evasion task at hand. 
As we will show in Sec.~\textcolor{red}{2.2}, directly applying spatial filtering to remove spatial fake patterns is ineffective. 
That is why in this work, we resort to a learning-based approach to perform implicit deep notch filtering in the spatial domain by first proactively breaking the fake patterns by adding overwhelming spatial noises (diminish the spikes in the image spectrum), and then applying per-pixel image deep filtering to remove the noise (\ie, restore image quality). We call this method \textbf{DeepNotch}.

\subsection{Deep Image Filtering for DeepNotch} \label{subsec:challenge}

Given a fake image $\mathbf{I}\in \mathds{R}^{H\times W}$ that is generated by some DeepFake technique (\eg, ProGAN), we aim to produce a new version of $\mathbf{I}$ (\ie, $\hat{\mathbf{I}}\in \mathds{R}^{H\times W}$) by reducing the artifact patterns introduced by that technique, retouching the input fake image to make it more realistic.
This task is significantly challenging since the content of fake images could be quite diverse and various GANs introduce different artifact textures into fake images. Because of the double complexity of the content and texture, it is rather difficult to employ DNNs to handle the image directly. 
%
To address the challenge, we formulate the {DeepNotch} as a general deep image filtering problem to perform implicit notch filtering, where the input image is processed by pixel-wise kernels estimated from an offline trained DNN:
\begin{align}
    \hat{\mathbf{I}} = \mathbf{K}\circledast\mathbf{I} \mathrm{~~~~with~~~~}\mathbf{K} = \text{DNN}(\mathbf{I}) \label{eq:imagefilter},
\end{align}
where $\text{DNN}(\cdot)$ denotes a UNet-like \cite{ronneberger2015u} deep neural network used to predict the pixel-wise kernels $\mathbf{K}\in \mathds{R}^{H\times W\times K^2}$. `$\circledast$' denotes pixel-wise filtering. More specifically, the $p$-th pixel in the image $\mathbf{I}$ is processed by the corresponding $p$-th kernel in $\mathbf{K}$ denoted as $\mathbf{K}_p\in \mathds{R}^{K\times K}$, where $K$ represents the kernel size. Then, we can offline train the DNN with fake-real image pairs and the $L_1$ loss function.
Obviously, the aforementioned structure seems a feasible and intuitive solution for DeepNotch: \emph{First}, the fake image is processed by only one single-layer filtering without any upsampling operations, avoiding the re-artifact risk. \emph{Second}, the kernels are generated from a DNN, which takes full advantage of deep learning in perceiving the image content and predicting suitable kernels for each pixel.

\emph{However, naively training such a DNN for DeepNotch, unfortunately, often cannot reduce the artifacts}. This wrong technical route is put in Fig.~\ref{fig:teaser} as a warning. As shown in Fig.~\ref{fig:teaser} (a), the typical artifact patterns of a DeepFake image can be visualized in both spatial and frequency domains. When employing the directly trained deep image filtering, \ie, Eq.~\eqref{eq:imagefilter}, to the image, the artifacts cannot be easily reduced, \ie, the trace indicated by red arrows in Fig.~\ref{fig:teaser} (b). The same conclusion is also reached on our large-scale evaluation (see Table~\ref{Table:GANFingerprint_Acc} of the experimental section), where the naive deep image filtering fails to retouch the fake image and misjudge DeepFake detectors. {The large-scale experimental results confirm that it is difficult to realize effective DeepNotch via single-layer filtering even if the DNN is equipped.} Furthermore, in theory, as indicated by the analysis of the spectral bias of Deep Neural Networks \cite{rahaman2019spectral}, DNNs tend to be biased toward learning lower-frequency functions. This bias implies that comprehending high-frequency variations (\eg, checkboard patterns) is a challenging task for DNNs. Consequently, it is not straightforward for DNNs to autonomously acquire the knowledge necessary to automatically learn notch filters capable of reducing artifacts in DeepFake images.

Such results force us to rethink the retouching solution. Intuitively, DNN-driven image filtering has demonstrated big advantages for image denoising. Meanwhile, noise can also be regarded as a perturbation in the spatial domain. 
Therefore, we come up with a bold idea, which first employs noise to destroy the artifacts, and then recovers the deliberate noisy image with DNN-driven image filtering. In other words, we find a way to implement notch filtering in the spatial domain implicitly, that is, first adding random noise to the image to break the periodic checkerboard-like noise pattern in the spatial domain, followed by per-pixel deep image filtering.  
We present how to realize this idea with the guidance of random noise (\ie, Sec.~\textcolor{red}{2.3}). Moreover, we further propose an advanced version by employing the popular adversarial attack to generate semantic-aware noise for guidance (\ie, Sec.~\textcolor{red}{2.4}).

\subsection{Random-Noise-Guided Image Filtering} \label{subsec:noise}
\begin{figure}
	\centering 
	\includegraphics[width=\columnwidth]{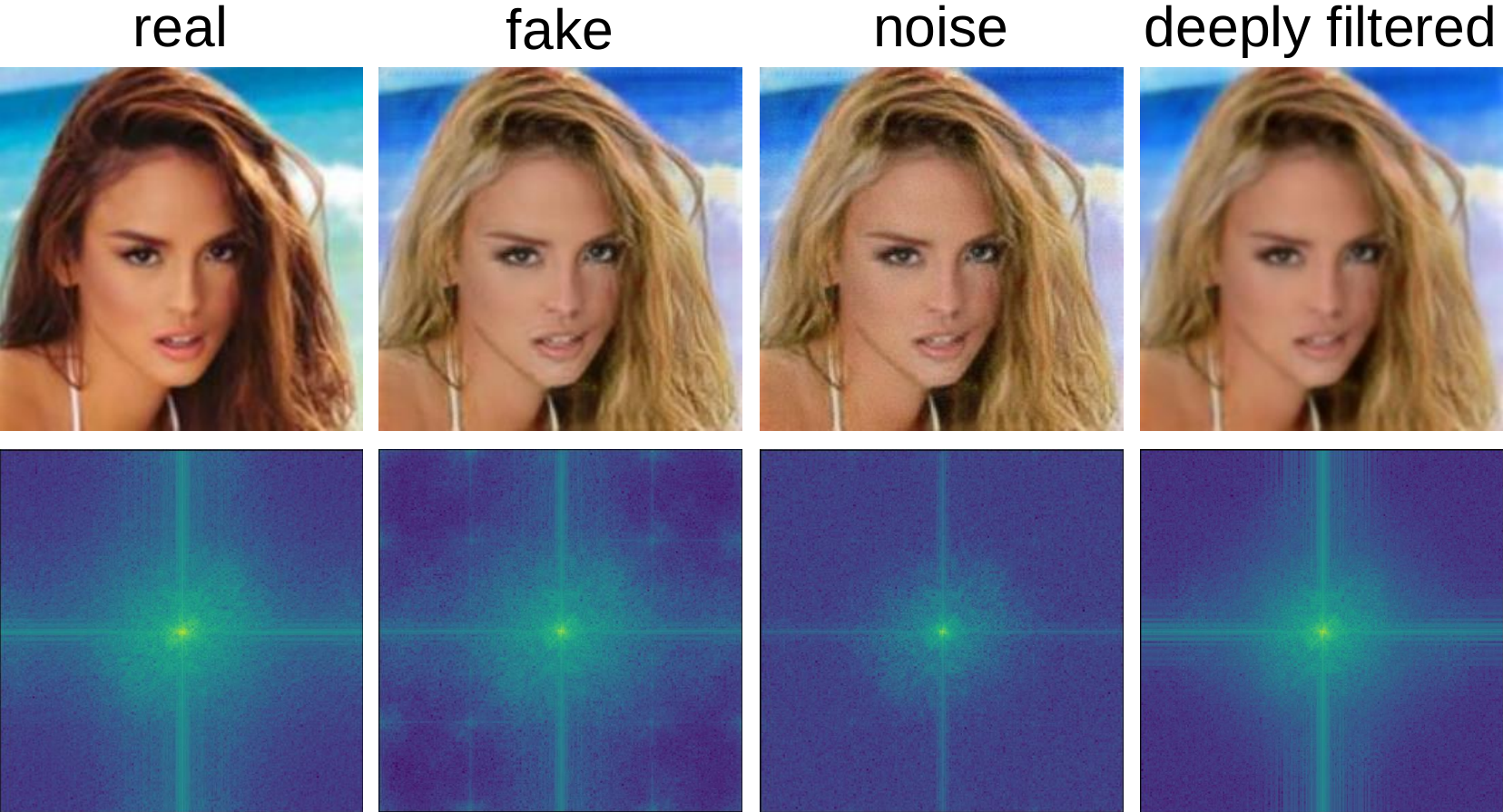}
	\caption{(L-R) Real, fake, noised, and deeply filtered image. Blobs are in the spectrum of the fake image selected from StarGAN. The noised image shows the result of adding Gaussian noise $(\sigma=5,\mu=0)$ to the fake image. It does not have blobs in the spectrum. In addition, the deeply filtered image using deep image filtering on the noised image also exhibits no artifact patterns.
	}
	\label{fig:toy_example}
\end{figure}

\begin{figure}[t]
    \centering 
    \includegraphics[width=\linewidth]{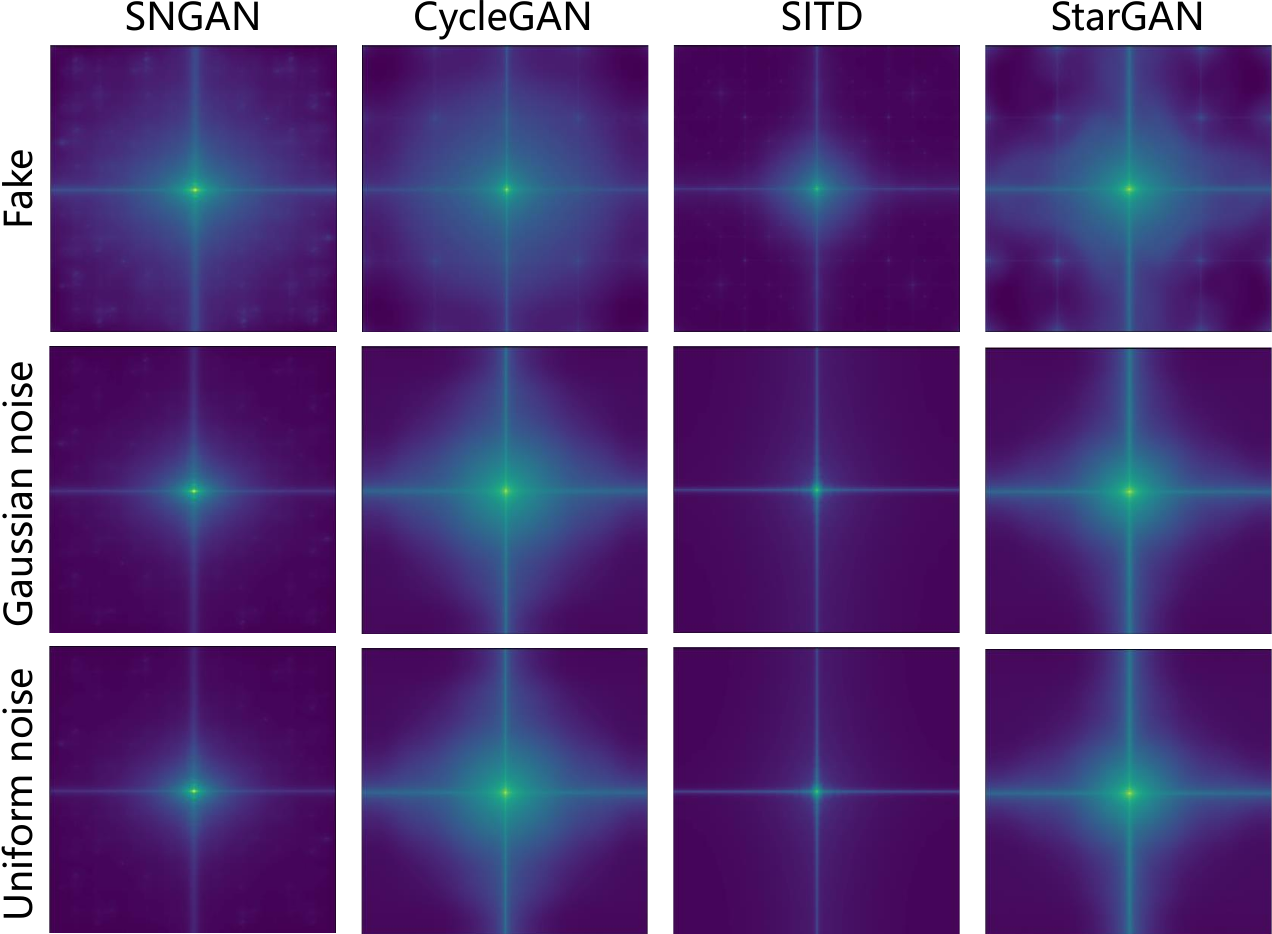}
    \caption{Average spectrums of DeepFake images and corresponding noised DeepFake images. We can find that noise can effectively reduce the artifacts in the spectrum.}
    \label{fig:spectrum_noise}
\end{figure}

We first give a simple example to explain that \textit{adding random noise can effectively reduce the artifacts}. As shown in Fig.~\ref{fig:toy_example}, we use a pair of real-fake images from StarGAN as an example. In the first two columns, we show real-fake image pairs and their corresponding spectrums. We can find that DeepFake techniques can corrupt the spectrum of the real face image and introduce the bright blob patterns shown in the spectrum of the fake image. After adding zero-mean Gaussian noise with a standard deviation of five to the fake image, see the third column, surprisingly, the artifacts in the spectrum are reduced. To demonstrate the effect of noise across different fake images, we conduct extra experiments by adding noise to DeepFake images and calculating the average spectrum of these images (compared with visualizing one image). As shown in Fig.~\ref{fig:spectrum_noise}, we illustrate the spectrums of SNGAN \cite{miyato2018spectral}, CycleGAN \cite{zhu2017unpaired}, SITD \cite{chen2018learning} and StarGAN \cite{choi2018stargan}. We choose these four GANs because the artifacts are obvious in their spectrums and thus we can display the effect of noise better. For each of them, we use dozens to tens of thousands (the same as in the experiment setting (Sec.~\ref{sec:setting})). The two noises chosen by us are Gaussian noise and uniform noise. The mean of Gaussian noise is 0 while the standard deviation is 10. The lower bound and upper bound of uniform noise are -20 and 20, respectively. Take the last column as an example (the StarGAN column), we can find that the average spectrum of StarGAN fake images has obvious artifacts (\ie, blobs). Compared with it, the spectrums (in Gaussian noise row and Uniform noise row) of the noised StarGAN fake image both almost have no artifacts. For other columns, we can achieve the same conclusion.

According to the above observation, we first propose the \textit{random-noise-guided image filtering} method. Given a fake image $\mathbf{I}$, we add random noise to it and process it with pixel-wise kernels, thus we can reformulate Eq.~\eqref{eq:imagefilter} as 
\begin{align}\label{eq:rdnoise_imagefilter}
    \hat{\mathbf{I}}=\mathbf{K}\circledast(\mathbf{I}+\mathbf{N}_\sigma),
\end{align}
where $\mathbf{N}_\sigma$ denotes the Gaussian noise map with standard deviation $\sigma$ and has the same size with $\mathbf{I}$. We can employ the same network in Eq.~\eqref{eq:imagefilter} to predict the kernels, which is offline trained with the fake-real image pairs and $L_1$ loss.

As shown in Fig.~\ref{fig:toy_example}, with the random-noise-guided image filtering, both the artifacts and deliberate noise are clearly reduced, \ie, the spectrum of the deeply filtered image is clean and similar to the real one. 
Nevertheless, as a kind of degradation, the random noise can potentially reduce the image quality.
Ideally, it is highly desirable the noise addition is as minor as possible, at the same time, reducing fake artifacts effectively.
As a result, the fake can not be detected by the advanced DeepFake detectors. This requires the noise addition method to be `smart' or semantic-aware, \ie, adding the noise to the proper locations.


\subsection{Adversarial-Noise-Guided Image Filtering} \label{subsec:advnoise}

Inspired by adversarial attacks, which perform semantic-aware and sparse noise perturbation, we further propose the \textit{adversarial-noise-guided image filtering}, where a simple DeepFake detection method trained by us is employed as a subject model to generate the adversarial noise and an $L_1$ constraint is used to reduce the noise strength.

Based on Eq.~\eqref{eq:rdnoise_imagefilter}, we retouch the fake image $\mathbf{I}$ via the guidance of an adversarial noise map and obtain 
\begin{align}\label{eq:advnoise_imagefilter}
    \hat{\mathbf{I}}=\mathbf{K}\circledast(\mathbf{I}+\mathbf{A}\odot\mathbf{N}_\sigma),
\end{align}
where `$\odot$' denotes the element-wise multiplication and $\mathbf{A}$ is a binary adversarial noise map with value `1' for adding noise $\mathbf{N}_\sigma$ to the fake image. 
It is highly desirable that $\mathbf{A}$ is sparse while pinpointing key positions where noises can do best to perturb the artifact patterns. To this end, we calculate $\mathbf{A}$ from the viewpoint of adversarial attack and employ a DeepFake detector (\ie, $\text{D}(\cdot)$) as a subject model. Please note that $\text{D}(\cdot)$ is a simple detector which unrelated to the detectors we aim to attack. Then, we have the following objective function:
\begin{align}\label{eq:advnoise_obj}
    \argmax_{\mathbf{A}}{J(\text{D}(\mathbf{I}+\mathbf{A}),y)+\|\mathbf{A}\|_1},
\end{align}
where $J(\cdot)$ denotes the cross-entropy loss function, $y$ is the ground-truth label of $\mathbf{I}$. Here, we have $y=1$ since $\mathbf{I}$ is a fake image. The second term encourages $\mathbf{A}$ to be sparse to add less noise to the fake image. 


In summary, algorithm~\ref{alg:DeepNotch} shows our method \textit{DeepNotch}. There are four key steps. First, in line 2, we select a fake image $\mathcal{I}$ from the fake image list $\{\mathcal{I}^{\mathrm{r}}\}$. Second, in line 3, we generate an adversarial guided map $\mathbf{A}$ produced by attacking a simple pre-trained DeepFake detector $\text{D}(\cdot)$ according to Eq.~\eqref{eq:advnoise_obj}. Third, in line 4, with a prepared noise map $\mathbf{N}_\sigma$ which is of uniform or Gaussian noise, we element-wise multiply $\mathbf{N}_\sigma$ with the adversarial guided map $\mathbf{A}$ to be $\mathbf{A}\odot\mathbf{N}_\sigma$ and add the noise to the fake image $\mathcal{I}$. At last, in line 5, the specified kernels $\mathbf{K}$ generated by deep neural network $\text{DNN}(\cdot)$ are efficiently used to embellish the noised image.
\begin{algorithm}[tb]
	{
		\caption{\small{DeepNotch}}\label{alg:DeepNotch}
		\KwIn{Fake images $\mathcal{I}^{\mathrm{r}}$, DeepFake detector $\text{D}(\cdot)$, Ground truth label $y$ of fake images, Cross-entropy loss function $J(\cdot)$, Noise map $\mathbf{N}_\sigma$, Pixel-wise kernels $\mathbf{K}$.}
		\KwOut{Reconstruction image $\mathbf{\hat{I}}$.}
 		\For{$i=1\ \mathrm{to}\ |\mathcal{I}^{\mathrm{r}}|$}{
 		    Sample an image $\mathbf{I}\sim\{\mathcal{I}^{\mathrm{r}}\}$\;
 		    Generate adversarial noise map $\mathbf{A}$ via  $\argmax_{\mathbf{A}}{J(\text{D}(\mathbf{I}+\mathbf{A}),y)+\|\mathbf{A}\|_1}$\; 
 		    Produce $\mathbf{I}^{'}$ by adding noise to image $\mathbf{I}$ via $\mathbf{I}^{'} = \mathbf{I}+\mathbf{A}\odot\mathbf{N}_\sigma$\;
 		    Apply image filtering to achieve reconstruction image $\mathbf{\hat{I}}$ from $\mathbf{I}^{'}$ via $\mathbf{\hat{I}} = \mathbf{K}\circledast \mathbf{I}^{'}$\;
    	}
	}
\end{algorithm}

\section{Experiments}\label{sec:exp}

In the experiments, we design two different validation methods to demonstrate the effectiveness of our method. First, we test whether the reconstructed images can reduce the detection accuracy of various fake image detection methods. Then, we further perform quantitative measurement of our reconstructed changing magnitude by using similarity metrics.

\subsection{Experimental Setup}\label{sec:setting}
\subsubsection{Fake detection methods} 
Existing fake detection methods largely fall into 3 categories. 
For each category, we choose one representative fake detection method. For fingerprint-based, image-based and spectrum-based methods, we select GANFingerprint \cite{yu2019attributing}, CNNDetector \cite{wang2020cnn}, and DCTA \cite{frank2020leveraging}, respectively. \textcolor{black}{\cite{yu2019attributing} is the latest work using fingerprint. CNNDetector \cite{wang2020cnn} detects a large number of GANs, which is suitable for testing the effectiveness of our method on different GANs. For spectrum-based methods, DCTA is popular and outstanding.}

\subsubsection{Datasets}
We tried our best to cover diverse datasets and select 3 popular real image datasets (CelebA \cite{liu2018large}, LSUN \cite{yu2015lsun}, FFHQ \cite{karras2019style}) used in previous work. CelebA and FFHQ are the most famous human face dataset while LSUN includes images of different rooms. 
We comprehensively tried a total of 16 GAN-based methods for fake image generation on the above real image datasets. Specifically, for GANFingerprint and DCTA, ProGAN \cite{karras2017progressive}, SNGAN \cite{miyato2018spectral}, CramerGAN \cite{bellemare2017cramer} and MMDGAN \cite{li2017mmd} are used as the fake image generators. For each of these four GAN-based image generation methods, we set the size of the testing dataset to be 10,000.
For CNNDetector, we select ProGAN, StyleGAN \cite{karras2019style}, BigGAN \cite{brock2018large}, CycleGAN \cite{zhu2017unpaired}, StarGAN \cite{choi2018stargan}, GauGAN \cite{park2019gaugan}, CRN \cite{shi2016end}, IMLE \cite{li2018implicit}, SITD \cite{chen2018learning}, SAN \cite{dai2019second}, DeepFakes \cite{rossler2019faceforensics++}, StyleGAN2 \cite{karras2020analyzing}, and Whichfaceisreal \cite{whichfaceisreal}, a total of 13 GAN-based image generation methods. The size of the testing dataset of these GAN-based image generation methods ranges from hundreds to thousands.
The elements in the datasets of CycleGAN, ProGAN, StyleGAN and StyleGAN2 have two or more categories (\eg, in StyleGAN2, four different categories (horse, car, cat, church) are contained). 
The datasets of other GANs have only one category. For example, Whichfaceisreal only has fake images of the human face.

\subsubsection{Evaluation settings}
In our experiment, we use KPN \cite{mildenhall2018burst} as the deep image filtering method. The training dataset of KPN has 10,000 pairs of images, each pair of which includes a real image chosen from CelebA and a fake image. The fake image is first produced by reconstructing the real image counterpart with STGAN \cite{liu2019stgan}, and then added with different types of noise. The two noises chosen by us are Gaussian noise and uniform noise. 
The mean of Gaussian noise is 0 while the standard deviation is 10. The lower bound and upper bound of uniform noise are -20 and 20, respectively. 
To add deliberate noise, we first train a simple DeepFake detector with ResNet50. Then, we use PGD \cite{madry2017towards} to adversarially attack this detector, on the fake images as the inputs to obtain adversarial guided maps. The epsilon (maximum perturbation for each pixel) is 0.04. At last, depending on the adversarial guided map, we choose locations of the fake images to add noise. Please note that our method is a post-processing black-box image reconstruction method that doesn't require any model information of the detectors. The DeepFake detector (based on Resnet50) used here is unrelated to the models of GANFingerprint, CNNDetector, and DCTA. 

\subsubsection{Metrics}
Detection accuracy is one of our main evaluation metrics. We compare the detection accuracy of fake images and reconstructed images for each method. In addition, to be comprehensive, we further use cosine similarity (COSS), peak signal-to-noise ratio (PSNR), and structural similarity (SSIM) to measure the similarity between a fake image and its reconstructed image counterpart. 
COSS is a common similarity metric that measures the cosine of the angle. We transform the RGB images to vectors before calculating COSS. PSNR is a widely used measurement for the reconstruction quality of lossy compression. SSIM is one of the most popular and useful metrics for measuring the similarity between two images. A large value of COSS, PSNR and SSIM, indicates a better result. The value ranges of COSS and SSIM are both in [0,1].

\begin{table*}
\centering
\caption{Detection accuracy before and after reconstruction of GAN-synthesized images in GANFingerprint. The accuracy in the columns with yellow background color is what we need to decrease. Because we use our method to reconstruct the GAN images of that column head.}
\resizebox{1\linewidth}{!}{
\begin{tabular}{l|l l l l l|l l l l l}
\toprule

\multirow{2}{*}{Accuracy(\%)} &  \multicolumn{5}{c|}{ProGAN (Pro)} & \multicolumn{5}{c}{SNGAN (SN)}   \\ 
                  & CelebA  & Pro  & SN &Cramer & MMD & CelebA  & Pro  & SN & Cramer & MMD \\ \midrule
Fake                & 0.03  & 99.91  & 0.01 & 0.03  &0.02 & 0.07  & 0.01  & 99.75 & 0.05 & 0.12  \\
BL-PCA   & 88.90 \textcolor{blue}{(+88.87)}  & \cellcolor{yellow!15}6.99   \textcolor{red}{(-92.92)}& 0.21 \textcolor{blue}{(+0.20)}& 0.07 \textcolor{blue}{(+0.04)}& 3.83 \textcolor{blue}{(+3.81)}& 46.10  \textcolor{blue}{(+46.03)}& 0.12  \textcolor{blue}{(+0.11)}& \cellcolor{yellow!15}50.86 \textcolor{red}{(-48.89)}& 0.16 \textcolor{blue}{(+0.11)}& 2.76 \textcolor{blue}{(+2.64)}\\ 
BL-KSVD &  21.50 \textcolor{blue}{(+21.47)}& \cellcolor{yellow!15}78.10  \textcolor{red}{(-21.81)}&0.10  \textcolor{blue}{(+0.09)}&0.20  \textcolor{blue}{(+0.17)}&0.10  \textcolor{blue}{(+0.08)}& 48.15  \textcolor{blue}{(+48.08)}& 4.60  \textcolor{blue}{(+4.59)}& \cellcolor{yellow!15}46.35 \textcolor{red}{(-53.40)}& 0.60 \textcolor{blue}{(+0.55)}& 0.30 \textcolor{blue}{(+0.18)}\\
StatAttack &  92.93 \textcolor{blue}{(+92.90)}& \cellcolor{yellow!15}4.79  \textcolor{red}{(-95.12)}& 0.27  \textcolor{blue}{(+0.26)} & 0.21  \textcolor{blue}{(+0.18)}& 1.80  \textcolor{blue}{(+1.78)}& 30.82 \textcolor{blue}{(+30.75)}& 0.06  \textcolor{blue}{(+0.05)}& \cellcolor{yellow!15}68.05 \textcolor{red}{(-31.70)}& 0.11 \textcolor{blue}{(+0.06)}& 0.96 \textcolor{blue}{(+0.84)}\\
DN(rn)-gau   & 93.39 \textcolor{blue}{(+93.36)}  & \cellcolor{yellow!15}4.77   \textcolor{red}{(-95.14)}& 0.14 \textcolor{blue}{(+0.13)}& 0.21 \textcolor{blue}{(+0.18)}& 1.49 \textcolor{blue}{(+1.47)}&  53.59 \textcolor{blue}{(+53.52)}&  0.05 \textcolor{blue}{(+0.04)}& \cellcolor{yellow!15}45.22 \textcolor{red}{(-54.53)}& 0.13 \textcolor{blue}{(+0.08)}& 1.01 \textcolor{blue}{(+0.89)}\\ 
DN(rn)-uni  &  91.79 \textcolor{blue}{(+91.76)}& \cellcolor{yellow!15}6.57  \textcolor{red}{(-93.34)}& 0.11  \textcolor{blue}{(+0.10)}& 0.20  \textcolor{blue}{(+0.17)}&0.22  \textcolor{blue}{(+0.20)}& 66.42  \textcolor{blue}{(+66.35)}& 0.16  \textcolor{blue}{(+0.15)}& \cellcolor{yellow!15}31.82 \textcolor{red}{(-67.93)}&  0.21 \textcolor{blue}{(+0.16)}& 1.39 \textcolor{blue}{(+1.27)}\\ 
DN(an)-gau  & \textbf{96.00}  \textcolor{blue}{(+95.97)}& \cellcolor{yellow!15}2.37  \textcolor{red}{(-97.54)}&  0.24 \textcolor{blue}{(+0.23)}& 0.19  \textcolor{blue}{(+0.16)}& 1.20 \textcolor{blue}{(+1.18)}&  62.89 \textcolor{blue}{(+62.82)}& 0.02  \textcolor{blue}{(+0.01)}& \cellcolor{yellow!15}36.24 \textcolor{red}{(-63.51)}& 0.06 \textcolor{blue}{(+0.01)}& 0.79 \textcolor{blue}{(+0.67)}\\ 
\textbf{DN(an)-uni}  & 95.76  \textcolor{blue}{(+95.73)}& \cellcolor{yellow!15}2.89  \textcolor{red}{(-97.02)}&  0.16 \textcolor{blue}{(+0.15)}& 0.12  \textcolor{blue}{(+0.09)}& 1.07 \textcolor{blue}{(+1.05)}&  \textbf{75.17} \textcolor{blue}{(+75.10)}& 0.07  \textcolor{blue}{(+0.06)}& \cellcolor{yellow!15}23.69 \textcolor{red}{(-76.07)}& 0.09 \textcolor{blue}{(+0.04)}& 0.98 \textcolor{blue}{(+0.86)}\\ 
\arrayrulecolor{gray}\hline
Fake-gau-5 & 92.75 \textcolor{blue}{(+92.72)} & 4.50 \cellcolor{yellow!15}\textcolor{red}{(-95.41)} & 0.17 \textcolor{blue}{(+0.16)} & 0.12 \textcolor{blue}{(+0.09)} & 2.46 \textcolor{blue}{(+2.44)} & 12.26 \textcolor{blue}{(+12.19)} & 0.00 \textcolor{red}{(-0.01)} & 87.15 \cellcolor{yellow!15}\textcolor{red}{(-12.60)}& 0.02 \textcolor{red}{(-0.03)} & 0.57 \textcolor{blue}{(+0.45)}\tabularnewline
Fake-gau-10   & 99.47 \textcolor{blue}{(+99.44)}  & \cellcolor{yellow!15}0.16 \textcolor{red}{(-99.75)}& 0.05 \textcolor{blue}{(+0.04)}& 0.06 \textcolor{blue}{(+0.03)}& 0.26 \textcolor{blue}{(+0.24)}&  78.52 \textcolor{blue}{(+78.45)}& 0.00  \textcolor{red}{(-0.01)}& 21.24\cellcolor{yellow!15} \textcolor{red}{(-78.51)}& 0.02 \textcolor{red}{(-0.03)}&0.22  \textcolor{blue}{(+0.10)}\\
Fake-uni-5 & 55.10 \textcolor{blue}{(+55.07)} & 36.42 \cellcolor{yellow!15}\textcolor{red}{(-63.49)} & 0.17 \textcolor{blue}{(+0.16)}& 0.52 \textcolor{blue}{(+0.49)} & 7.79 \textcolor{blue}{(+7.77)}& 0.66 \textcolor{blue}{(+0.59)}& 0.00 \textcolor{red}{(-0.01)} & 99.05 \cellcolor{yellow!15}\textcolor{red}{(-0.70)} & 0.04 \textcolor{red}{(-0.01)}& 0.25 \textcolor{blue}{(+0.13)}\tabularnewline
Fake-uni-10 & 95.96 \textcolor{blue}{(+95.93)} & 2.20 \cellcolor{yellow!15}\textcolor{red}{(-97.71)} & 0.06 \textcolor{blue}{(+0.05)} & 0.08 \textcolor{blue}{(+0.05)}& 1.70 \textcolor{blue}{(+1.68)}& 23.15 \textcolor{blue}{(+23.08)}& 0.00 \textcolor{red}{(-0.01)}& 76.33 \cellcolor{yellow!15}\textcolor{red}{(-23.42)}& 0.01 \textcolor{red}{(-0.04)}& 0.51 \textcolor{blue}{(+0.39)}\tabularnewline
Fake-uni-15 & 99.24 \textcolor{blue}{(+99.21)}& 0.33 \cellcolor{yellow!15}\textcolor{red}{(-99.58)} & 0.06 \textcolor{blue}{(+0.05)}& 0.05 \textcolor{blue}{(+0.02)}& 0.32 \textcolor{blue}{(+0.30)}& 65.75 \textcolor{blue}{(+65.68)}& 0.00 \textcolor{red}{(-0.01)}& 34.09 \cellcolor{yellow!15}\textcolor{red}{(-65.66)}& 0.00 \textcolor{red}{(-0.05)}& 0.16 \textcolor{blue}{(+0.04)}\tabularnewline
Fake-uni-20   & 99.66 \textcolor{blue}{(+99.63)}  & 0.07\cellcolor{yellow!15}   \textcolor{red}{(-99.84)}& 0.04 \textcolor{blue}{(+0.03)}& 0.04 \textcolor{blue}{(+0.01)}& 0.19 \textcolor{blue}{(+0.17)}&  87.41 \textcolor{blue}{(+87.34)}& 0.00  \textcolor{red}{(-0.01)}&12.48 \cellcolor{yellow!15}\textcolor{red}{(-87.27)}& 0.02 \textcolor{red}{(-0.03)}& 0.09 \textcolor{red}{(-0.03)}\\ 
Filt(nn)   & 0.03 \textcolor{blue}{(0)}  &  \cellcolor{yellow!15}99.91  \textcolor{blue}{(0)}& 0.01 \textcolor{blue}{(0)}& 0.03 \textcolor{blue}{(0)}&  0.02 \textcolor{blue}{(0)}&  0.95 \textcolor{blue}{(+0.88)}&   0.62 \textcolor{blue}{(+0.61)}& \cellcolor{yellow!15}97.06 \textcolor{red}{(-2.69)}& 0.93 \textcolor{blue}{(+0.88)}& 0.44 \textcolor{blue}{(+0.32)} \\ 
\arrayrulecolor{black}\midrule  

\multirow{2}{*}{Accuracy(\%)} &  \multicolumn{5}{c|}{CramerGAN (Cramer)} & \multicolumn{5}{c}{MMDGAN (MMD)}    \\                     & CelebA  & Pro  & SN &Cramer & MMD & CelebA  & Pro  & SN & Cramer & MMD \\ \arrayrulecolor{black}\midrule

Fake                & 0.00  & 0.02  & 0.02 & 99.76 &  0.20 & 0.11  & 0.01  & 0.04 & 0.27 & 99.57 \\
BL-PCA   & 54.85  \textcolor{blue}{(+54.85)}& 0.35 \textcolor{blue}{(+0.33)}& 0.93 \textcolor{blue}{(+0.91)}& \cellcolor{yellow!15}35.07 \textcolor{red}{(-64.69)}& 8.80 \textcolor{blue}{(+8.60)}& 45.94  \textcolor{blue}{(+45.83)}& 0.13  \textcolor{blue}{(+0.12)}& 0.20 \textcolor{blue}{(+0.16)}& 0.03 \textcolor{red}{(-0.24)}& \cellcolor{yellow!15}53.70 \textcolor{red}{(-45.87)}\\ 
BL-KSVD  &  28.70 \textcolor{blue}{(+28.70)}& 14.90 \textcolor{blue}{(+14.88)} & 0.10 \textcolor{blue}{(+0.08)}& \cellcolor{yellow!15}55.60 \textcolor{red}{(-44.16)}& 0.70 \textcolor{blue}{(+0.50)}&  47.40 \textcolor{blue}{(+47.29)}& 14.20   \textcolor{blue}{(+14.19)}& 0.30 \textcolor{blue}{(+0.26)}& 0.70 \textcolor{blue}{(+0.43)}& \cellcolor{yellow!15}37.40 \textcolor{red}{(-62.17)}\\
StatAttack &  61.71 \textcolor{blue}{(+61.71)}& 0.10  \textcolor{blue}{(+0.08)}& 1.32  \textcolor{blue}{(+1.30)}& 28.86  \cellcolor{yellow!15}\textcolor{red}{(-70.90)}& 8.01  \textcolor{blue}{(+7.99)}& 67.93  \textcolor{blue}{(+67.82)}& 0.08  \textcolor{blue}{(+0.07)}& 0.62 \textcolor{blue}{(+0.58)}& 0.25 \textcolor{red}{(-0.02)}& 31.12 \cellcolor{yellow!15}\textcolor{red}{(-68.45)}\\
DN(rn)-gau & 66.18 \textcolor{blue}{(+66.18)}& 0.29 \textcolor{blue}{(+0.27)}& 0.71 \textcolor{blue}{(+0.69)}& \cellcolor{yellow!15}29.71 \textcolor{red}{(-70.05)}& 3.11 \textcolor{blue}{(+2.91)}&  69.96 \textcolor{blue}{(+69.85)}& 0.10 \textcolor{blue}{(+0.09)}& 0.24  \textcolor{blue}{(+0.20)}& 0.15 \textcolor{red}{(-0.12)}& \cellcolor{yellow!15}29.55 \textcolor{red}{(-70.02)}\\
DN(rn)-uni  &  70.13 \textcolor{blue}{(+70.13)}& 0.55 \textcolor{blue}{(+0.53)} & 0.51 \textcolor{blue}{(+0.49)}& \cellcolor{yellow!15}25.89 \textcolor{red}{(-73.87)}& 2.92 \textcolor{blue}{(+2.72)}& 74.00  \textcolor{blue}{(+73.89)}&  0.20  \textcolor{blue}{(+0.19)}& 0.30 \textcolor{blue}{(+0.26)}& 0.10 \textcolor{red}{(-0.17)}& \cellcolor{yellow!15}25.40 \textcolor{red}{(-74.43)}\\
DN(an)-gau  & 77.80  \textcolor{blue}{(+77.80)}& 0.11\textcolor{blue}{(+0.09)} & 0.71 \textcolor{blue}{(+0.69)}& \cellcolor{yellow!15}18.43 \textcolor{red}{(-81.33)}& 2.95 \textcolor{blue}{(+2.75)}&  80.15 \textcolor{blue}{(+80.04)}&  0.05  \textcolor{blue}{(+0.04)}& 0.24 \textcolor{blue}{(+0.20)}& 0.04 \textcolor{red}{(-0.23)}& \cellcolor{yellow!15}19.52 \textcolor{red}{(-80.05)}\\
\textbf{DN(an)-uni}  & \textbf{81.54}  \textcolor{blue}{(+81.54)}& 0.16 \textcolor{blue}{(+0.14)} & 0.56 \textcolor{blue}{(+0.54)}& \cellcolor{yellow!15}15.22 \textcolor{red}{(-84.54)}& 2.52 \textcolor{blue}{(+2.32)}&  \textbf{83.42} \textcolor{blue}{(+83.31)}&  0.09  \textcolor{blue}{(+0.08)}& 0.17 \textcolor{blue}{(+0.13)}& 0.04 \textcolor{red}{(-0.23)}& \cellcolor{yellow!15}16.28 \textcolor{red}{(-83.29)}\\
\arrayrulecolor{gray}\hline
Fake-gau-5 & 41.37 \textcolor{blue}{(+41.37)}& 0.01 \textcolor{red}{(-0.01)}& 0.78 \textcolor{blue}{(+0.76)}& 51.15 \cellcolor{yellow!15}\textcolor{red}{(-48.61)}& 6.69 \textcolor{blue}{(+6.49)}& 45.54 \textcolor{blue}{(+45.43)}& 0.00 \textcolor{red}{(-0.01)}& 0.12 \textcolor{blue}{(+0.08)}& 0.02 \textcolor{red}{(-0.25)}& 54.32 \cellcolor{yellow!15}\textcolor{red}{(-45.25)}\tabularnewline
Fake-gau-10   & 92.95 \textcolor{blue}{(+92.95)}  & 0.00   \textcolor{red}{(-0.02)}&  0.31 \textcolor{blue}{(+0.29)}& 5.17 \cellcolor{yellow!15}\textcolor{red}{(-94.59)}& 1.57  \textcolor{blue}{(+1.37)}&  94.30 \textcolor{blue}{(+94.19)}& 0.00  \textcolor{red}{(-0.01)}& 0.11  \textcolor{blue}{(+0.07)}& 0.03 \textcolor{red}{(-0.24)}& 5.56 \cellcolor{yellow!15}\textcolor{red}{(-94.01)}\\
Fake-uni-5 & 2.28 \textcolor{blue}{(+2.28)}& 0.03 \textcolor{blue}{(+0.01)}& 0.25 \textcolor{blue}{(+0.23)}& 94.29 \cellcolor{yellow!15}\textcolor{red}{(-5.47)}& 3.15 \textcolor{blue}{(+2.95)}& 4.86 \textcolor{blue}{(+4.75)}& 0.00 \textcolor{red}{(-0.01)}& 0.12 \textcolor{blue}{(+0.08)}& 0.03 \textcolor{red}{(-0.24)}& 94.99 \cellcolor{yellow!15}\textcolor{red}{(-4.58)}\tabularnewline
Fake-uni-10 & 58.53 \textcolor{blue}{(+58.53)}& 0.01 \textcolor{red}{(-0.01)}& 0.63 \textcolor{blue}{(+0.61)}& 35.04 \cellcolor{yellow!15}\textcolor{red}{(-64.72)}& 5.79 \textcolor{blue}{(+5.59)}& 62.05 \textcolor{blue}{(+61.94)}& 0.00 \textcolor{red}{(-0.11)}& 0.25 \textcolor{blue}{(+0.21)}& 0.02 \textcolor{red}{(-0.25)}& 37.68 \cellcolor{yellow!15}\textcolor{red}{(-61.89)}\tabularnewline
Fake-uni-15 & 89.00 \textcolor{blue}{(+89.00)}& 0.01 \textcolor{red}{(-0.01)}& 0.38 \textcolor{blue}{(+0.36)}& 8.15 \cellcolor{yellow!15}\textcolor{red}{(-91.61)}& 2.46 \textcolor{blue}{(+2.26)}& 90.44 \textcolor{blue}{(+90.33)}& 0.00 \textcolor{red}{(-0.01)}& 0.12 \textcolor{blue}{(+0.08)}& 0.01 \textcolor{red}{(-0.26)}& 9.43 \cellcolor{yellow!15}\textcolor{red}{(-90.14)}\tabularnewline
Fake-uni-20 &  95.51 \textcolor{blue}{(+95.51)}  & 0.00    \textcolor{red}{(-0.02)}& 0.19 \textcolor{blue}{(+0.17)}& 3.32 \cellcolor{yellow!15}\textcolor{red}{(-96.44)}& 0.98 \textcolor{blue}{(+0.78)}&  96.76 \textcolor{blue}{(+96.65)}& 0.00  \textcolor{red}{(-0.01)}& 0.06 \textcolor{blue}{(+0.02)}&  0.01 \textcolor{red}{(-0.26)}& 3.17 \cellcolor{yellow!15}\textcolor{red}{(-96.40)}\\ 
Filt(nn)   & 0.01 \textcolor{blue}{(+0.01)}  & 0.11   \textcolor{blue}{(+0.09)}& 0.03 \textcolor{blue}{(+0.01)}&  \cellcolor{yellow!15}99.69 \textcolor{red}{(-0.07)}& 0.16 \textcolor{red}{(-0.04)}&  0.23 \textcolor{blue}{(+0.12)}&  0.11 \textcolor{blue}{(+0.10)}& 0.11 \textcolor{blue}{(+0.07)}& 1.46 \textcolor{blue}{(+1.19)}& \cellcolor{yellow!15}98.09 \textcolor{red}{(-1.48)}\\ 
\arrayrulecolor{black}\bottomrule
\end{tabular}}
\label{Table:GANFingerprint_Acc}
\end{table*}

\subsection{Experiment I: Evading GANFingerprint}\label{sec:GANF}

\begin{table}
\centering
\caption{Detection accuracy before and after reconstruction of CelebA images in GANFingerprint.}
\resizebox{1\linewidth}{!}{
\begin{tabular}{l|l l l l l}
\toprule

{Accuracy(\%)} & CelebA  & ProGAN  & SNGAN &CramerGAN & MMDGAN \\ \midrule
Real                     & 96.17  & 3.76  & 0.03 & 0.01  &0.03 \\
Real-DN(rn)-gau          & 96.96  & 2.21  & 0.40 & 0.06  &0.38 \\
Real-DN(rn)-uni          & 97.07  & 2.29  & 0.34 & 0.04  &0.26 \\

\arrayrulecolor{black}\bottomrule
\end{tabular}}
\label{Table:GANFingerprint_real_Acc}
\end{table}

GANFingerprint can judge which GAN-based image generation method is used to produce the fake image.
For a double-check and validation of our settings, we reproduce their experiments on fake image detection. For each GAN-based fake image generation method (\ie, ProGAN, SNGAN, CramerGAN, MMDGAN), we produce 10,000 fake images from 10,000 randomly chosen CelebA real images. Furthermore, we replace their fake images with our reconstructed images to test the detection accuracy of their method.

\begin{table}
\centering
\caption{Detection accuracy of fake and reconstructed images (\ie, DN(rn)-uni) in retrained GANFingerprint model. The model is retrained with reconstructed images of DN(rn)-uni.}
\resizebox{1\linewidth}{!}{
\begin{tabular}{l|l l l l l|l l l l l}
\toprule

\multirow{2}{*}{Accuracy(\%)} &  \multicolumn{5}{c|}{ProGAN (Pro)} & \multicolumn{5}{c}{SNGAN (SN)}   \\ 
                  & CelebA  & Pro  & SN &Cramer & MMD & CelebA  & Pro  & SN & Cramer & MMD \\ \midrule
Fake & 0.03 & 99.71 & 0.03 & 0.17 & 0.06 & 0.00 & 0.24 & 99.19 & 0.37 & 0.20\tabularnewline
DN(rn)-uni & 0.09 & 96.11 & 0.35 & 2.18 & 1.27 & 0.04 & 2.48 & 89.20 & 4.06 & 4.22\tabularnewline
\midrule  

\multirow{2}{*}{Accuracy(\%)} &  \multicolumn{5}{c|}{CramerGAN (Cramer)} & \multicolumn{5}{c}{MMDGAN (MMD)}    \\                     & CelebA  & Pro  & SN &Cramer & MMD & CelebA  & Pro  & SN & Cramer & MMD \\ \arrayrulecolor{black}\midrule
Fake & 0.01 & 0.14 & 0.03 & 99.63 & 0.19 & 0.01 & 0.54 & 0.17 & 0.91 & 98.37\tabularnewline
DN(rn)-uni & 0.07 & 1.47 & 0.57 & 95.58 & 2.31 & 0.11 & 3.40 & 1.49 & 5.62 & 89.38\tabularnewline

\arrayrulecolor{black}\bottomrule
\end{tabular}}
\label{Table:GANFingerprint_Retrain_Acc}
\end{table}

\subsubsection{Compare with the baseline}
Table \ref{Table:GANFingerprint_Acc} shows the detection accuracy of different GANs. We use ProGAN (Pro) in the second column as an example for elaboration. In particular, Pro has five sub-items (columns): CelebA, ProGAN (Pro), SNGAN (SN), CramerGAN (Cramer), MMDGAN (MMD). These sub-items represent the possibilities of the ProGAN images to be classified as one of them. In the first column are the types of the source of input images.

\begin{table}
\centering
\caption{Detection accuracy before and after reconstruction of GAN-synthesized images in DCTA.}
\resizebox{1\linewidth}{!}{
\begin{tabular}{l|c c|c c}
\toprule

\multirow{2}{*}{Accuracy(\%)} &  \multicolumn{2}{c|}{ProGAN} & \multicolumn{2}{c}{SNGAN}   \\ 
                  & CelebA  & Fake & CelebA  & Fake\\ \midrule
Fake           &    0.43&99.57 & 0.22& 99.78  \\
BL-PCA  & 74.50 \textcolor{blue}{(+74.07)}& 25.50 \textcolor{red}{(-74.07)}& 56.85 \textcolor{blue}{(+56.63)}& 43.15 \textcolor{red}{(-56.63)} \\
BL-KSVD & 23.40 \textcolor{blue}{(+23.13)} & 76.60 \textcolor{red}{(-23.13)}& 71.60 \textcolor{blue}{(+70.92)}& 28.40 \textcolor{red}{(-70.92)}\\
StatAttack & 76.18 \textcolor{blue}{(+75.75)}& 23.82 \textcolor{red}{(-75.75)}& 76.06  \textcolor{blue}{(+75.85)}& 23.94 \textcolor{red}{(-75.84)}\\
DN(rn)-gau  & 82.07 \textcolor{blue}{(+81.64)}& 17.93 \textcolor{red}{(-81.64)}& 56.30 \textcolor{blue}{(+56.08)}& 43.70 \textcolor{red}{(-56.08)}\\ 
DN(rn)-uni  & 83.84 \textcolor{blue}{(+83.41)}& 16.16 \textcolor{red}{(-83.41)}& 70.11 \textcolor{blue}{(+69.89)}& 29.89 \textcolor{red}{(-69.89)}\\ 
DN(an)-gau  & 87.59 \textcolor{blue}{(+87.16)}& 12.41 \textcolor{red}{(-87.16)}& 60.43 \textcolor{blue}{(+60.21)} & 39.57  \textcolor{red}{(-60.21)}\\
DN(an)-uni  &88.55 \textcolor{blue}{(+88.12)}& 11.45 \textcolor{red}{(-88.12)}& 73.62 \textcolor{blue}{(+73.40)}& 26.38 \textcolor{red}{(-73.40)} \\ 
\arrayrulecolor{black}\midrule  

\multirow{2}{*}{Accuracy(\%)} &  \multicolumn{2}{c|}{CramerGAN} & \multicolumn{2}{c}{MMDGAN}    \\                     & CelebA  & Fake & CelebA  & Fake \\ \arrayrulecolor{black}\midrule

Fake      & 0.27 &99.73 &0.68 & 99.32\\
BL-PCA   & 30.22 \textcolor{blue}{(+29.95)}& 69.78 \textcolor{red}{(-29.95)}& 24.71 \textcolor{blue}{(+24.03)}& 75.29 \textcolor{red}{(-24.03)}\\ 
BL-KSVD  & 23.40 \textcolor{blue}{(+23.13)} & 76.60 \textcolor{red}{(-23.13)}& 71.60 \textcolor{blue}{(+70.92)}& 28.40 \textcolor{red}{(-70.92)}\\
StatAttack  & 75.99 \textcolor{blue}{(+75.72)} & 23.94 \textcolor{red}{(-75.72)}& 76.01 \textcolor{blue}{(+75.33)}& 23.99 \textcolor{red}{(-75.33)}\\
DN(rn)-gau  & 43.66 \textcolor{blue}{(+43.39)} & 56.34 \textcolor{red}{(-43.39)}& 45.01 \textcolor{blue}{(+44.33)}& 54.99  \textcolor{red}{(-44.33)}\\
DN(rn)-uni  & 50.86 \textcolor{blue}{(+50.59)} & 49.14 \textcolor{red}{(-50.59)}& 51.98 \textcolor{blue}{(+51.30)}& 48.02 \textcolor{red}{(-51.30)} \\
DN(an)-gau  & 49.93 \textcolor{blue}{(+49.66)}& 50.07 \textcolor{red}{(-49.66)}& 50.30 \textcolor{blue}{(+49.62)}& 49.70 \textcolor{red}{(-49.62)}\\
DN(an)-uni & 57.29 \textcolor{blue}{(+57.02)}& 42.71 \textcolor{red}{(-57.02)}& 56.69 \textcolor{blue}{(+56.01)}&43.31 \textcolor{red}{(-56.01)} \\
\arrayrulecolor{black}\bottomrule
\end{tabular}}
\label{Table:DCTA_Acc}
\end{table}
The baseline is the state-of-the-art method \cite{huang2020fakepolisher}. They use the shallow reconstruction method to polish fake images. In the seven rows above the gray line, Fake represents the fake images. BL-PCA and BL-KSVD represent the PCA-based and KSVD-based reconstruction of the baseline, respectively. DN(rn)-gau, DN(rn)-uni, DN(an)-gau, DN(an)-uni mean the reconstructed images generated from our method. In particular, `rn' and `an' mean adding random noise and adversarial noise respectively. `uni' and `gau' mean adding uniform noise and Gaussian noise respectively.

In the Fake row, the input images are 10,000 ProGAN fake images. We can find that 99.91\% of the fake images have been considered as being produced by Pro. The percentages of the ProGAN fake images that are misclassified as CelebA, SN, Cramer and MMD are 0.03\%, 0.01\%, 0.03\% and 0.02\%. In the table, we highlight the difference in detection accuracy between reconstructed images and fake images (\eg, by color and number). As shown in the DN(an)-uni row, the inputs are the reconstructed images by adding and denoising adversarial-noise-guided uniform noise on the counterpart fake images. Most of the 10,000 reconstructed images are considered as CelebA type. The accuracy raises from 0.03\% to 95.76\%. We use blue color and \textcolor{blue}{(+95.73)} to highlight the difference. Similarly, the ratio of images classified into Pro decreases from 99.91\% to 2.89\%. We use red color and \textcolor{red}{(-97.02)} to show the difference. Most of the fake images generated by ProGAN are misclassified to be real images after using our method. Compared with BL-PCA and BL-KSVD, DN(rn)-gau and DN(rn)-uni degrade the detector more. 
Our method also surpasses the BL significantly in the other three parts (\ie, SN, Cramer, MMD).

\subsubsection{Ablation study}\label{sec:ablation_study}
To verify the effectiveness of key technology: adding noise, deep image filtering and adversarial guided map.

\begin{itemize}[leftmargin=*]
\item \textit{Adding noise.} We also take ProGAN as an example. The conclusion is the same on the other three GANs. The experimental results are below the gray line. In the first column, Fake-gau-10 and Fake-uni-20 represent only adding Gaussian noise (gau) or uniform noise (uni) to the Fake images. The accuracy of Fake-gau-10 and Fake-uni-20 images being classified as CelebA type raises from 0.03\% to 99.47\% and 99.66\% respectively. This phenomenon shows the effectiveness of noise in reducing artifact patterns. On the other hand, if we only use deep image filtering without adding noise to the fake image, the result is shown in the row of Filt(nn), which means the reconstructed images with no noise (nn) added. We can find that the classification accuracy is basically unchanged. Furthermore, we evaluate the effect of different levels of noise on misleading the detector. The mean of Gaussian noise is 0 while the standard deviation is 5 and 10 (Fake-gau-5 and Fake-gau-10). The lower bounds and upper bounds of uniform noise are $\pm$5 (Fake-uni-5), $\pm$10 (Fake-uni-10), $\pm$15 (Fake-uni-15), $\pm$20 (Fake-uni-20), respectively. We can find that, with the increase in noise level, the noised fake images are more likely to evade GANFingerprint. At the same time, as shown in Fig.~\ref{fig:spectrum_noise_level}, the more noise added, the fewer artifacts in the spectrum. This evidence supports that reducing artifacts can help to evade the DeepFake detector. We also conduct similar experiments in Table~\ref{Table:CNNDetector_Acc} and achieve the same conclusion. 

\begin{figure*}[]
    \centering 
    \includegraphics[width=\linewidth]{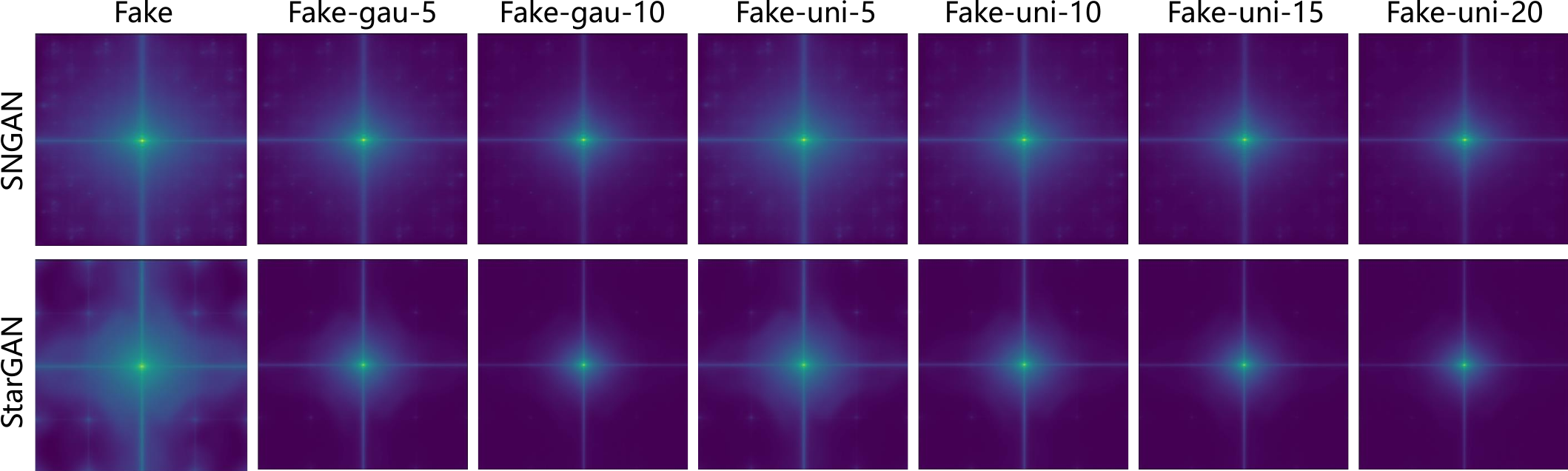}
    \caption{Spectrums of SNGAN and StarGAN images under different levels of noises. The more noise added, the fewer artifacts in the spectrum.}
    \label{fig:spectrum_noise_level}
\end{figure*}

\item \textit{Deep image filtering.} We add guided filter \cite{he2010guided} and BM3D \cite{dabov2006image} to compare the effect of other classical denoising methods with deep image filtering. Among the four GANs, to evade GANFingerprint, the accuracy of fake images being classified as real images with the guided filter or BM3D is 19.86\% and 1.35\% lower than using the KPN filter with kernel size three. From the experiment, we can find that our method surpasses the guided filter by far and the performance of BM3D is similar to deep filtering. However, the time of generating 40,000 images upon experiment takes BM3D 24 hours while our method only needs half an hour.

\item \textit{Adversarial guided map.} Here we show the effect of the adversarial guided map by comparing DN(rn)-uni and DN(an)-uni. Since the guided noise only uses partial regions of the fake image to add noise, we increased the intensity of the noise to match the total tensity of DN(rn)-uni. In Table \ref{Table:GANFingerprint_Acc}, DN(an)-uni uses 80\% of the area to add uniform noise, thus the lower and upper bounds of uniform noise are 1.25 times (-25/25) of that in DN(rn)-uni (-20/20). We highlight the best performance of the reconstruction method of all four GANs in bold font. The results of DN(an)-uni are much better than DN(rn)-gau and DN(rn)-uni. Similarly, The results of DN(an)-gau are much better than DN(rn)-gau and DN(rn)-uni too. We can also find that DN(an)-uni is a little better than DN(an)-gau.
\end{itemize}

Furthermore, we apply our method on real images to see whether the method actually transforms the images to real image distribution or it only creates a new variant of image distribution that GANFingerprint has not seen yet. As shown in Table \ref{Table:GANFingerprint_real_Acc}, Real means using real images as input. Real-DN(rn)-gau and Real-DN(rn)-uni means the reconstructed images are produced by random-noise-guided DeepNotch with Gaussian and uniform noise on real images respectively. The accuracies of reconstructed images being classified as CelebA are higher than that of real images. The fact that the real images \emph{stay classified} as real and almost all the DeepFake images \emph{become classified} as real, is a testimony that our method is actually shifting the DeepFake distribution towards the real-image distribution.

We also retrain the GANFingerprint model with reconstructed images to test whether the retrained model can detect reconstructed images. As shown in Table \ref{Table:GANFingerprint_Retrain_Acc}, the model is retrained with reconstructed images generated from DN(rn)-uni. The model can successfully classify reconstructed images as fake, which means the images reconstructed by our method can help to improve the DeepFake detector to a new version.

\subsection{Experiment II: Evading DCTA}\label{sec:DCTA}

DCTA has the same testing dataset as GANFingerprint. We follow the exact evaluation setting in its original paper, where the spectrums of the images are used as inputs. 
For each category of CelebA, ProGAN, SNGAN, CramerGAN and MMDGAN, we use 9,600 fake images as the input. The testing dataset has a total of 48,000 images.
As shown in Table \ref{Table:DCTA_Acc}, DCTA successfully detects fake images with high accuracy. However, on reconstructed images, it decreases significantly. DN(rn)-gau, DN(rn)-uni, DN(an)-gau, DN(an)-uni all successfully drop the classification accuracy of DCTA and do better than baselines. DCTA only obtains 22.59\% accuracy on DN(rn)-gau reconstructed images, at a dramatic drop of 66.4\%. For DN(rn)-uni reconstruction, it shows slightly better (21.73\%) than DN(rn)-gau reconstruction. Furthermore, DN(an)-uni and DN(an)-gau, the methods which exploit adversarial-noise-guided image filtering, achieve better performance than their corresponding random-noise-guided counterpart respectively. DN(an)-uni does the best, which is slightly better than DN(an)-gau. The results demonstrate the effectiveness of reconstructed images in misleading the DCTA.

\begin{table*}[htb]
\centering
\caption{Detection accuracy before and after reconstruction of GAN-synthesized images in CNNDetector.}
\resizebox{1\linewidth}{!}{
\begin{tabular}{l l l l l l l l l l l l l l l}
\toprule

\multicolumn{2}{c}{Accuracy(\%)}& ProGAN & StyleGAN & BigGAN & CycleGAN & StarGAN & GauGAN & CRN & IMLE & SITD & SAN & DeepFakes & StyleGAN2 & Whichfaceisreal \\ 
 \toprule
\multirow{14}{*}{\rotatebox{90}{prob0.1}} & 
 Fake   & 99.9 & 74.2 & 46.8 & 78.8 & 86.7 & 64.8 & 99.8 & 99.8 & 86.7 & 1.83 & 6.86 & 68.8 & 74.3 \\ 
&  BL-PCA   & 42.3 \textcolor{red}{(-57.6)} & 3.90 \textcolor{red}{(-70.3)} & 12.3 \textcolor{red}{(-34.5)} & 35.8 \textcolor{red}{(-43.0)} & 36.0 \textcolor{red}{(-50.7)} & 14.2 \textcolor{red}{(-50.6)} & 6.50 \textcolor{red}{(-93.3)}  & 19.4 \textcolor{red}{(-80.4)} & 3.89 \textcolor{red}{(-82.81)} & 3.20 \textcolor{blue}{(+1.37)} & 1.33 \textcolor{red}{(-5.53)} & 11.9 \textcolor{red}{(-56.9)} & 1.40 \textcolor{red}{(-72.9)}  \\ 
  &  BL-KSVD  & 94.9 \textcolor{red}{(-5.0)} & 33.7 \textcolor{red}{(-40.5)} & 30.0 \textcolor{red}{(-16.8)} & 68.7 \textcolor{red}{(-10.1)} & 48.0 \textcolor{red}{(-38.7)} & 51.0 \textcolor{red}{(-13.8)} & 79.0 \textcolor{red}{(-20.8)} & 88.0 \textcolor{red}{(-11.8)} & 45.0 \textcolor{red}{(-41.7)} & 8.00 \textcolor{blue}{(+6.17)} & 0.00 \textcolor{red}{(-6.86)} & 33.5 \textcolor{red}{(-35.3)} & 50.0 \textcolor{red}{(-24.3)} \\
     & StatAttack & 0.09 \textcolor{red}{(-99.81)} & 0.00 \textcolor{red}{(-74.20)} & 0.00 \textcolor{red}{(-46.80)} & 0.02 \textcolor{red}{(-78.78)} & 0.02 \textcolor{red}{(-86.68)} & 0.02 \textcolor{red}{(-64.78)} & 0.00 \textcolor{red}{(-99.8)} & 0.00 \textcolor{red}{(-99.8)} & 0.00 \textcolor{red}{(-86.7)} & 0.00 \textcolor{red}{(-1.83)} & 0.03 \textcolor{red}{(-6.83)} & 0.00 \textcolor{red}{(-68.8)} & 0.00 \textcolor{red}{(-74.3)}\tabularnewline
  &  DN(rn)-gau   & 96.8 \textcolor{red}{(-3.1)} & 28.6 \textcolor{red}{(-45.6)} & 25.0 \textcolor{red}{(-21.8)} & 61.8 \textcolor{red}{(-17.0)} & 43.2 \textcolor{red}{(-43.5)} & 36.9 \textcolor{red}{(-27.9)} &  10.5 \textcolor{red}{(-89.3)}  & 20.2 \textcolor{red}{(-79.6)} & 86.7 \textcolor{red}{(0)} & 7.76 \textcolor{blue}{(+5.93)} & 3.04  \textcolor{red}{(-3.82)} & 37.0 \textcolor{red}{(-31.8)} & 2.90  \textcolor{red}{(-71.4)}  \\ 
  &  DN(rn)-uni   & 95.3 \textcolor{red}{(-4.6)} & 19.9 \textcolor{red}{(-54.3)} & 21.7 \textcolor{red}{(-25.1)} &  60.9 \textcolor{red}{(-17.9)} & 33.8 \textcolor{red}{(-52.9)} & 29.9 \textcolor{red}{(-34.9)} & 4.95  \textcolor{red}{(-34.9)}  & 15.8 \textcolor{red}{(-84.0)} & 85.6 \textcolor{red}{(-1.10)} & 6.85 \textcolor{blue}{(+5.02)} & 1.82 \textcolor{red}{(-5.04)} & 25.3 \textcolor{red}{(-43.5)} &  4.10 \textcolor{red}{(-70.2)} \\ 
  
&  DN(an)-gau  & 66.8 \textcolor{red}{(-33.1)}& 12.5 \textcolor{red}{(-61.7)}  & 5.05 \textcolor{red}{(-41.75)} & 18.8 \textcolor{red}{(-60.0)} & 14.3 \textcolor{red}{(-72.4)} & 5.38 \textcolor{red}{(-59.42)} & 0.16 \textcolor{red}{(-99.64)} &0.83 \textcolor{red}{(-98.97)} & 3.33 \textcolor{red}{(-83.37)} & 3.65 \textcolor{blue}{(+1.82)} & 0.56 \textcolor{red}{(-6.3)} &7.35 \textcolor{red}{(-61.45)} & 21.5 \textcolor{red}{(-52.8)} \\
 &  DN(an)-uni  & 59.5 \textcolor{red}{(-40.4)} & 14.9 \textcolor{red}{(-59.3)} & 4.05 \textcolor{red}{(-42.75)}& 15.6 \textcolor{red}{(-63.2)} & 9.70 \textcolor{red}{(-77.0)}& 3.64 \textcolor{red}{(-61.16)}  &0.27 \textcolor{red}{(-99.53)} & 1.80 \textcolor{red}{(-98.0)}& 0.00 \textcolor{red}{(-86.7)} & 5.48 \textcolor{blue}{(+3.65)}& 0.89 \textcolor{red}{(-5.97)} & 9.69 \textcolor{red}{(-59.11)}& 20.7 \textcolor{red}{(-53.6)} \\ 
  & Fake-gau-5 & 91.5 \textcolor{red}{(-8.40)} & 19.7 \textcolor{red}{(-54.50)} & 11.1 \textcolor{red}{(-35.70)} & 30.0 \textcolor{red}{(-48.80)} & 28.2 \textcolor{red}{(-58.50)} & 8.06 \textcolor{red}{(-56.74)} & 0.32 \textcolor{red}{(-99.48)} & 1.36 \textcolor{red}{(-98.44)} & 6.66 \textcolor{red}{(-80.04)} & 2.28 \textcolor{blue}{(+0.45)} & 0.03 \textcolor{red}{(-6.83)} & 22.0 \textcolor{red}{(-46.80)} & 0.10 \textcolor{red}{(-74.20)}\tabularnewline
 & Fake-gau-10 & 32.1 \textcolor{red}{(-67.80)} & 1.53 \textcolor{red}{(-72.67)} & 2.25 \textcolor{red}{(-44.55)} & 3.48 \textcolor{red}{(-75.32)} & 0.55 \textcolor{red}{(-86.15)} & 0.68 \textcolor{red}{(-64.12)} & 0.00 \textcolor{red}{(-99.80)} & 0.00 \textcolor{red}{(-99.80)} & 0.55 \textcolor{red}{(-86.15)} & 0.00 \textcolor{red}{(-1.83)} & 0.00 \textcolor{red}{(-6.86)} & 0.45 \textcolor{red}{(-68.35)} & 0.00 \textcolor{red}{(-74.30)}\tabularnewline
 & Fake-uni-5 & 99.9 \textcolor{red}{(0.00)} & 57.4 \textcolor{red}{(-16.80)} & 28.0 \textcolor{red}{(-18.80)} & 64.4 \textcolor{red}{(-14.40)} & 77.5 \textcolor{red}{(-9.20)} & 35.1 \textcolor{red}{(-29.70)} & 45.1 \textcolor{red}{(-54.70)} & 61.7 \textcolor{red}{(-38.10)} & 39.4 \textcolor{red}{(-47.30)} & 7.76 \textcolor{blue}{(+5.93)} & 10.0 \textcolor{blue}{(+3.14)} & 76.3 \textcolor{blue}{(+7.50)} & 8.70 \textcolor{red}{(-65.60)}\tabularnewline
 & Fake-uni-10 & 84.3 \textcolor{red}{(-15.60)} & 12.3 \textcolor{red}{(-61.90)} & 7.60 \textcolor{red}{(-39.20)} & 20.5 \textcolor{red}{(-58.30)} & 14.1 \textcolor{red}{(-72.60)} & 4.84 \textcolor{red}{(-59.96)} & 0.04 \textcolor{red}{(-99.76)} & 0.21 \textcolor{red}{(-99.59)} & 3.33 \textcolor{red}{(-83.37)} & 1.36 \textcolor{red}{(-0.47)} & 0.00 \textcolor{red}{(-6.86)} & 11.3 \textcolor{red}{(-57.49)} & 0.00 \textcolor{red}{(-74.30)}\tabularnewline
 & Fake-uni-15 & 44.6 \textcolor{red}{(-55.30)} & 2.40 \textcolor{red}{(-71.80)} & 2.85 \textcolor{red}{(-43.95)} & 6.73 \textcolor{red}{(-72.07)} & 1.55 \textcolor{red}{(-85.15)} & 0.96 \textcolor{red}{(-63.84)} & 0.00 \textcolor{red}{(-99.80)} & 0.01 \textcolor{red}{(-99.79)} & 1.11 \textcolor{red}{(-85.59)} & 0.00 \textcolor{red}{(-1.83)} & 0.00 \textcolor{red}{(-6.86)} & 1.08 \textcolor{red}{(-67.72)} & 0.00 \textcolor{red}{(-74.30)}\tabularnewline
 & Fake-uni-20 & 20.8 \textcolor{red}{(-79.10)} & 0.93 \textcolor{red}{(-73.27)} & 1.60 \textcolor{red}{(-45.20)} & 2.42 \textcolor{red}{(-76.38)} & 0.15 \textcolor{red}{(-86.55)} & 0.36 \textcolor{red}{(-64.44)} & 0.00 \textcolor{red}{(-99.80)} & 0.00 \textcolor{red}{(-99.80)} & 0.55 \textcolor{red}{(-86.15)} & 0.00 \textcolor{red}{(-1.83)} & 0.00 \textcolor{red}{(-6.86)} & 0.18 \textcolor{red}{(-68.62)} & 0.00 \textcolor{red}{(-74.30)}\tabularnewline \hline
 \multirow{14}{*}{\rotatebox{90}{prob0.5}} & 
Fake   & 100 & 46.9 & 18.9 & 62.9 & 62.7 & 59.2 & 76.0 & 88.9 & 63.9 & 0.00 & 2.52 & 36.9 & 28.6   \\ 
&  
 BL-PCA   &  71.6 \textcolor{red}{(-28.4)} & 3.00 \textcolor{red}{(-43.9)} & 6.45 \textcolor{red}{(-12.45)} & 30.9 \textcolor{red}{(-32.0)} & 42.1 \textcolor{red}{(-20.6)} & 22.8 \textcolor{red}{(-36.4)} & 4.36 \textcolor{red}{(-71.64)} & 16.7 \textcolor{red}{(-72.2)} & 1.12 \textcolor{red}{(-62.78)} & 0.00 \textcolor{red}{(0)} & 1.89 \textcolor{red}{(-0.61)} & 6.84 \textcolor{red}{(-30.06)} & 0.70 \textcolor{red}{(-27.9)} \\ 
  &  BL-KSVD  & 96.7 \textcolor{red}{(-3.30)} & 20.7 \textcolor{red}{(-26.2)} & 9.00 \textcolor{red}{(-9.90)} & 44.2 \textcolor{red}{(-18.7)} & 37.0 \textcolor{red}{(-25.7)} & 44.0 \textcolor{red}{(-15.2)} & 22.0 \textcolor{red}{(-54.0)} & 60.0 \textcolor{red}{(-28.9)} & 36.0 \textcolor{red}{(-27.9)} & 0.00 \textcolor{red}{(0)} & 2.00 \textcolor{red}{(-0.52)} & 13.0 \textcolor{red}{(-23.9)} & 18.0 \textcolor{red}{(-10.6)} \\
  & StatAttack & 0.14 \textcolor{red}{(-99.86)} & 0.00 \textcolor{red}{(-46.9)} & 0.00 \textcolor{red}{(-18.9)} & 0.02 \textcolor{red}{(-62.88)} & 0.02 \textcolor{red}{(-62.68)} & 0.03 \textcolor{red}{(-59.17)} & 0.00 \textcolor{red}{(-76.0)} & 0.01 \textcolor{red}{(-88.89)} & 0.01 \textcolor{red}{(-63.89)} & 0.00 \textcolor{red}{(0)} & 0.01 \textcolor{red}{(-2.51)} & 0.00 \textcolor{red}{(-36.9)} & 0.00 \textcolor{red}{(-28.6)}\tabularnewline
   &  DN(rn)-gau    & 98.5 \textcolor{red}{(-1.5)} &  17.0 \textcolor{red}{(-29.9)} & 9.15 \textcolor{red}{(-9.75)} & 41.9 \textcolor{red}{(-21.0)} & 27.5 \textcolor{red}{(-35.2)} & 37.5 \textcolor{red}{(-21.7)} &  2.16 \textcolor{red}{(-73.84)}  & 8.95 \textcolor{red}{(-79.95)} & 86.1 \textcolor{blue}{(+22.2)} & 1.37 \textcolor{blue}{(+1.37)} &  2.82 \textcolor{blue}{(+0.30)} & 15.0 \textcolor{red}{(-21.9)} & 7.90  \textcolor{red}{(-20.7)}  \\  
  &  DN(rn)-uni   & 97.9 \textcolor{red}{(-2.1)} & 14.2 \textcolor{red}{(-32.7)} & 9.10 \textcolor{red}{(-9.8)} & 41.9 \textcolor{red}{(-21.0)} & 25.0 \textcolor{red}{(-37.7)} & 34.5 \textcolor{red}{(-24.7)} & 1.55 \textcolor{red}{(-74.45)}  & 8.02 \textcolor{red}{(-80.88)} & 83.3 \textcolor{blue}{(+19.4)} & 2.28 \textcolor{blue}{(+2.28)} & 2.48 \textcolor{red}{(-0.04)} & 13.3 \textcolor{red}{(-23.6)} &  9.70 \textcolor{red}{(-18.9)}  \\ 
 &  DN(an)-gau     & 77.4 \textcolor{red}{(-22.6)} & 9.05 \textcolor{red}{(-37.85)}& 3.20 \textcolor{red}{(-15.7)}& 16.8 \textcolor{red}{(-46.1)}& 12.7 \textcolor{red}{(-50.0)}& 9.00 \textcolor{red}{(-50.2)}& 0.02 \textcolor{red}{(-75.98)} & 0.27 \textcolor{red}{(-88.63)}& 0.56 \textcolor{red}{(-63.3)}& 0.00 \textcolor{red}{(0)}&2.97 \textcolor{blue}{(+0.45)}& 3.61 \textcolor{red}{(-33.29)}& 12.1 \textcolor{red}{(-16.5)}\\  
  &  DN(an)-uni    & 71.2 \textcolor{red}{(-28.8)}& 11.0 \textcolor{red}{(-35.9)}& 15.7 \textcolor{red}{(-3.2)}& 2.60 \textcolor{red}{(-60.3)}& 9.91 \textcolor{red}{(-52.79)} & 6.84 \textcolor{red}{(-52.36)}& 0.00 \textcolor{red}{(-76.0)}& 0.36 \textcolor{red}{(-88.5)}& 0.00 \textcolor{red}{(-63.9)} & 1.37 \textcolor{blue}{(+1.37)}& 2.67 \textcolor{blue}{(+0.15)} & 4.42 \textcolor{red}{(-32.48)} & 21.8 \textcolor{red}{(-6.8)}\\
   & Fake-gau-5 & 96.7 \textcolor{red}{(-3.30)} & 15.4 \textcolor{red}{(-31.50)} & 7.35 \textcolor{red}{(-11.55)} & 38.3 \textcolor{red}{(-24.60)} & 37.1 \textcolor{red}{(-25.60)} & 20.4 \textcolor{red}{(-38.80)} & 0.21 \textcolor{red}{(-75.79)} & 1.36 \textcolor{red}{(-87.54)} & 8.33 \textcolor{red}{(-55.57)} & 0.45 \textcolor{blue}{(+0.45)} & 2.26 \textcolor{red}{(-0.26)} & 9.83 \textcolor{red}{(-27.07)} & 1.10 \textcolor{red}{(-27.50)}\tabularnewline
 & Fake-gau-10 & 57.3 \textcolor{red}{(-42.70)} & 2.73 \textcolor{red}{(-44.17)} & 2.80 \textcolor{red}{(-16.10)} & 14.6 \textcolor{red}{(-48.30)} & 5.75 \textcolor{red}{(-56.95)} & 3.70 \textcolor{red}{(-55.50)} & 0.00 \textcolor{red}{(-76.00)} & 0.00 \textcolor{red}{(-88.90)} & 2.22 \textcolor{red}{(-61.68)} & 0.00 \textcolor{red}{(0.00)} & 0.92 \textcolor{red}{(-1.60)} & 0.45 \textcolor{red}{(-36.45)} & 0.10 \textcolor{red}{(-28.50)}\tabularnewline
 & Fake-uni-5 & 100 \textcolor{red}{(0.00)} & 36.5 \textcolor{red}{(-10.40)} & 13.3 \textcolor{red}{(-5.60)} & 54.9 \textcolor{red}{(-8.00)} & 58.0 \textcolor{red}{(-4.70)} & 43.2 \textcolor{red}{(-16.00)} & 15.1 \textcolor{red}{(-60.90)} & 36.3 \textcolor{red}{(-52.60)} & 44.4 \textcolor{red}{(-19.50)} & 1.36 \textcolor{blue}{(+1.36)} & 4.85 \textcolor{blue}{(+2.33)} & 35.4 \textcolor{red}{(-1.50)} & 6.20 \textcolor{red}{(-22.40)}\tabularnewline
 & Fake-uni-10 & 93.3 \textcolor{red}{(-6.70)} & 10.6 \textcolor{red}{(-36.30)} & 5.85 \textcolor{red}{(-13.05)} & 32 \textcolor{red}{(-30.90)} & 28.3 \textcolor{red}{(-34.40)} & 14.4 \textcolor{red}{(-44.80)} & 0.01 \textcolor{red}{(-75.99)} & 0.37 \textcolor{red}{(-88.53)} & 5.00 \textcolor{red}{(-58.90)} & 0.00 \textcolor{red}{(0.00)} & 1.77 \textcolor{red}{(-0.75)} & 4.94 \textcolor{red}{(-31.96)} & 1.10 \textcolor{red}{(-27.50)}\tabularnewline
 & Fake-uni-15 & 68.4 \textcolor{red}{(-31.60)} & 3.53 \textcolor{red}{(-43.37)} & 3.40 \textcolor{red}{(-15.50)} & 18.2 \textcolor{red}{(-44.70)} & 8.80 \textcolor{red}{(-53.90)} & 5.36 \textcolor{red}{(-53.84)} & 0.00 \textcolor{red}{(-76.00)} & 0.00 \textcolor{red}{(-88.90)} & 1.66 \textcolor{red}{(-62.24)} & 0.00 \textcolor{red}{(0.00)} & 1.03 \textcolor{red}{(-1.49)} & 1.01 \textcolor{red}{(-35.89)} & 0.20 \textcolor{red}{(-28.40)}\tabularnewline
 & Fake-uni-20 & 44.8 \textcolor{red}{(-55.20)} & 1.88 \textcolor{red}{(-45.02)} & 2.15 \textcolor{red}{(-16.75)} & 9.46 \textcolor{red}{(-53.44)} & 3.20 \textcolor{red}{(-59.50)} & 2.68 \textcolor{red}{(-56.52)} & 0.00 \textcolor{red}{(-76.00)} & 0.00 \textcolor{red}{(-88.90)} & 1.11 \textcolor{red}{(-62.79)} & 0.00 \textcolor{red}{(0.00)} & 0.63 \textcolor{red}{(-1.89)} & 0.22 \textcolor{red}{(-36.68)} & 0.10 \textcolor{red}{(-28.50)}\tabularnewline
\bottomrule
\end{tabular}
}
\label{Table:CNNDetector_Acc}
\end{table*}
\subsection{Experiment III: Evading CNNDetector}\label{sec:CNNDetector}
We further perform a large-scale evaluation on CNNDetector with a total of 13 GAN-based image generation methods. The testing dataset of CNNDetector contains diverse types of images (\eg, animals, human faces, road). For each GAN category, the size of the testing dataset ranges from hundreds to thousands.

As shown in the first column of Table \ref{Table:CNNDetector_Acc}, the two models used by CNNDetector are prob0.1 and prob0.5. We can find that both prob0.1 and prob0.5 have achieved high accuracy on a large proportion of GANs. Here we take the case prob0.1 as an example to introduce the table. In the second row, Fake is the testing dataset that contains fake images. The experimental result on fake images is the same as that in CNNDetector.

DN(rn)-gau, DN(rn)-uni, DN(an)-gau, DN(an)-uni in the 5-8 rows are the results of our method. The accuracy data with red decrements are where our method succeeds. Compared with the detection accuracy of fake images, DN(rn)-gau and DN(rn)-uni reconstructed images both drop the accuracy of CNNDetector. Furthermore, DN(an)-gau and DN(an)-uni drop the accuracy of CNNDetector model more and have comparable performance with baselines.
\begin{figure}[htb]
\centering
\setlength{\abovecaptionskip}{0.2cm}   
\setlength{\belowcaptionskip}{-0.5cm}   
\subfigure[\scriptsize{ProGAN}]{
\includegraphics[width=0.45\columnwidth]{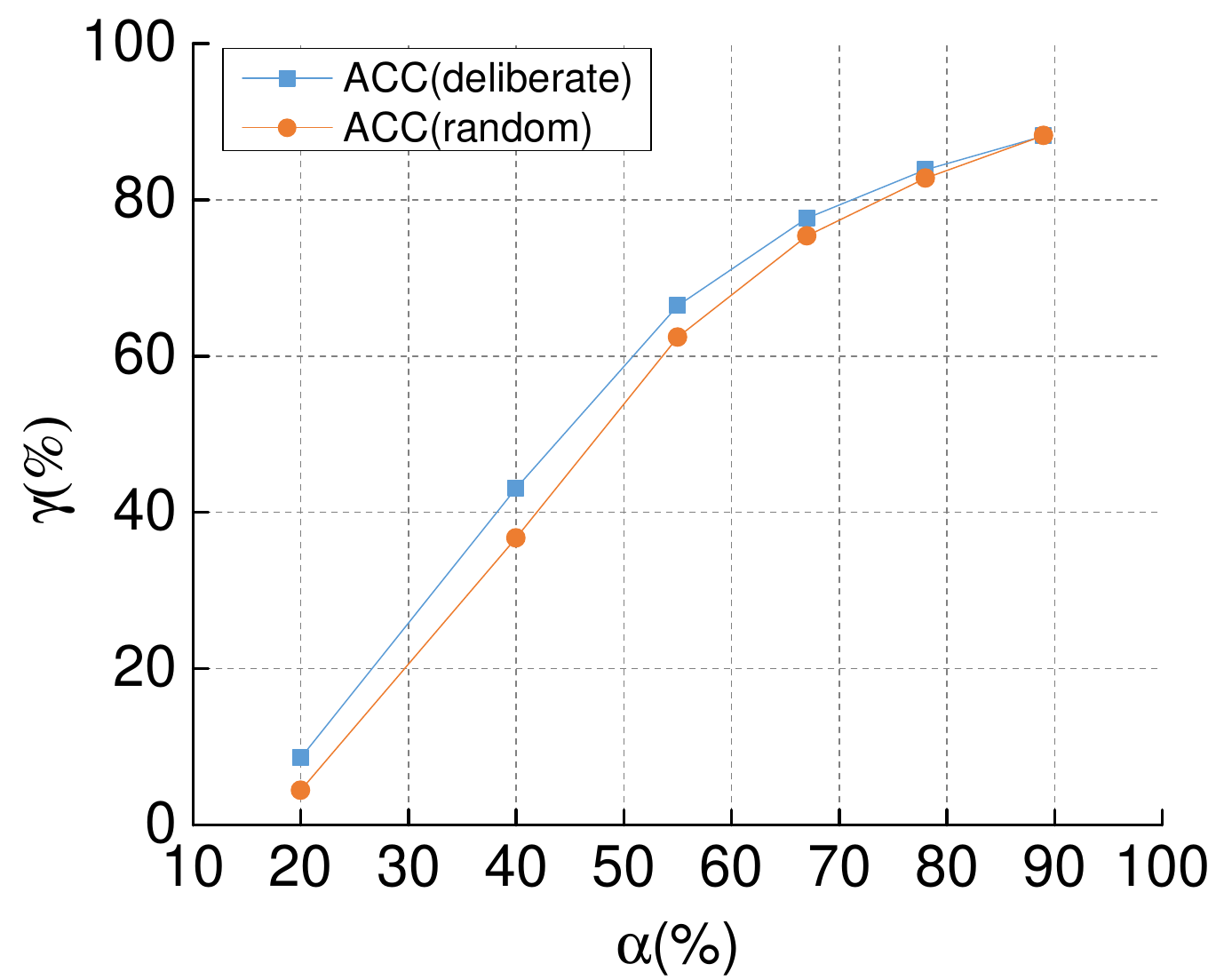}
}
\subfigure[\scriptsize{SNGAN}]{
\includegraphics[width=0.45\columnwidth]{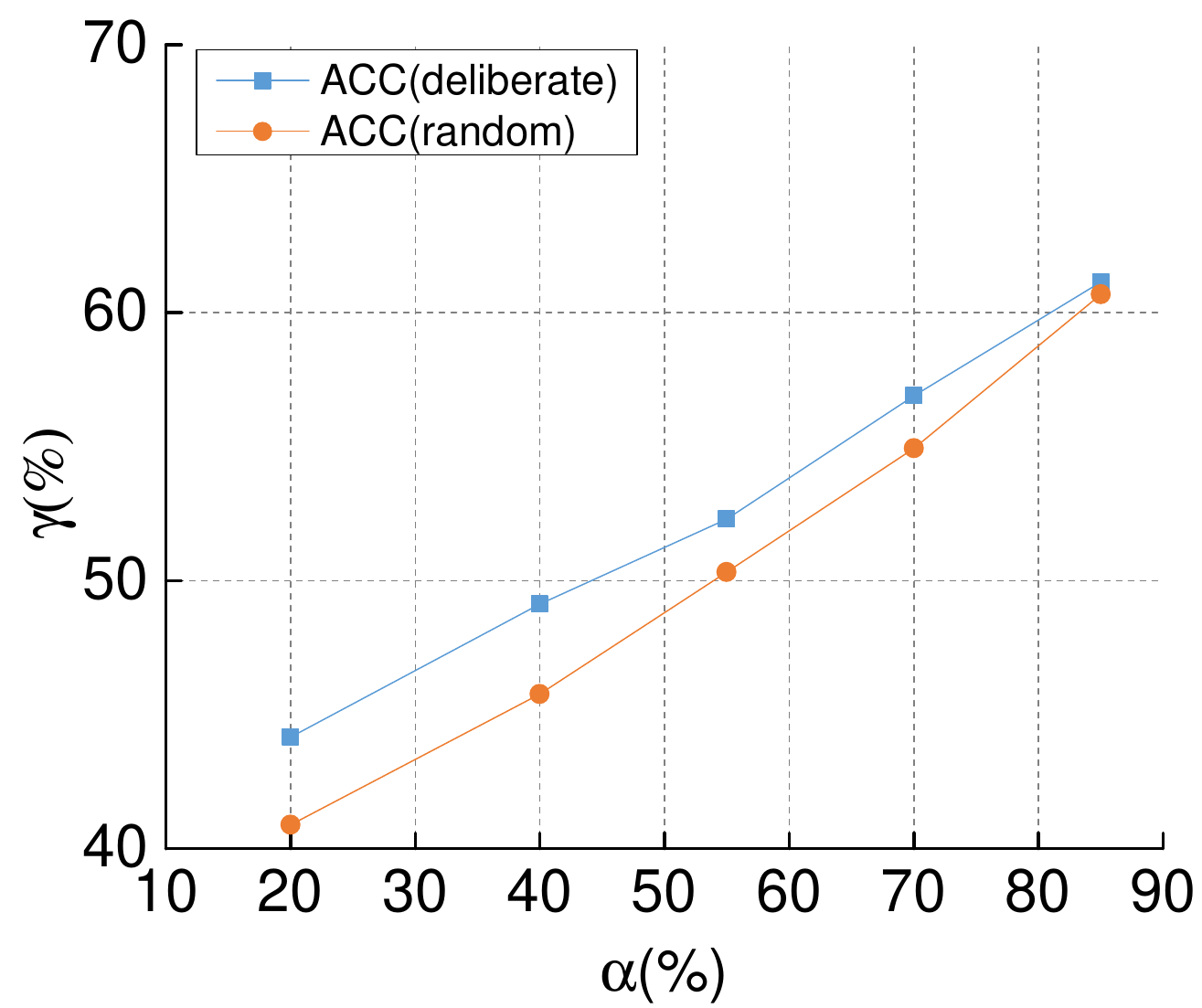}
}
\subfigure[\scriptsize{CramerGAN}]{
\includegraphics[width=0.45\columnwidth]{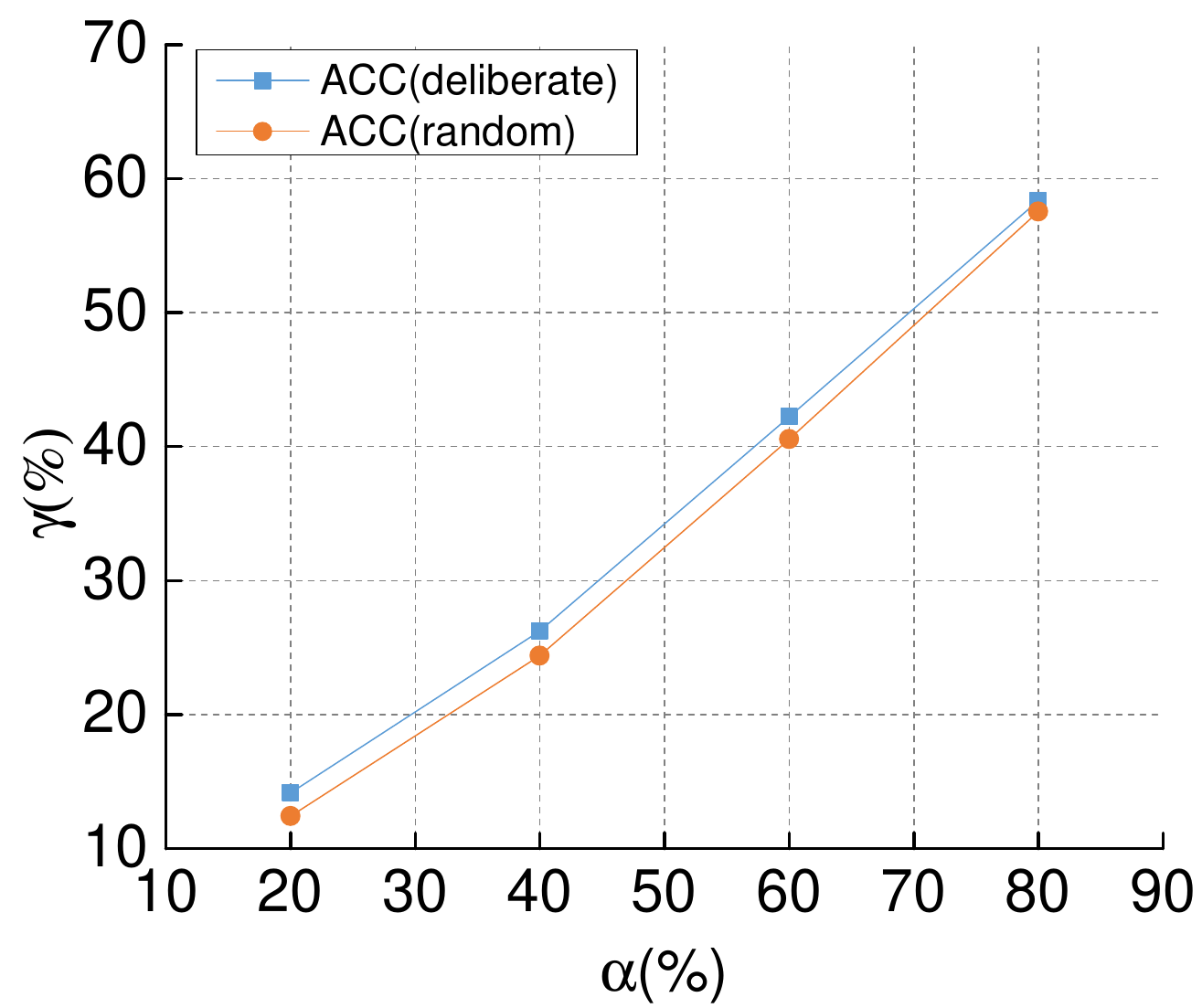}
}
\subfigure[\scriptsize{MMDGAN}]{
\includegraphics[width=0.45\columnwidth]{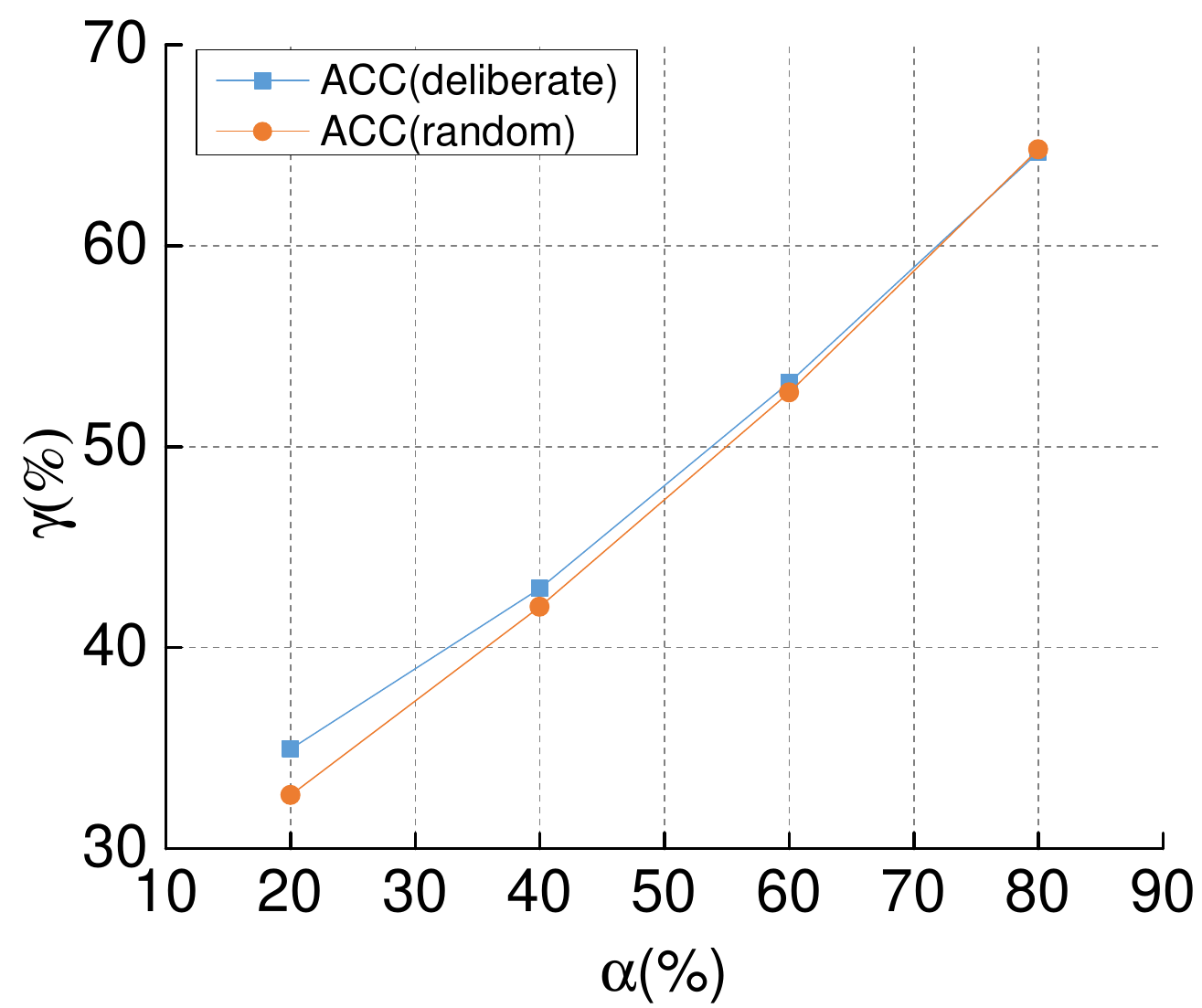}
}
\caption{Comparison between deliberate noise and random noise. $\alpha$ represents the percentage of the region in the image where noise is added. $\gamma$ means the possibility of reconstructed images being classified as CelebA types by method GANFingerprint.}
\label{fig:region_compare}
\end{figure}

\begin{figure*}[htb]
\centering
\includegraphics[scale=0.3]{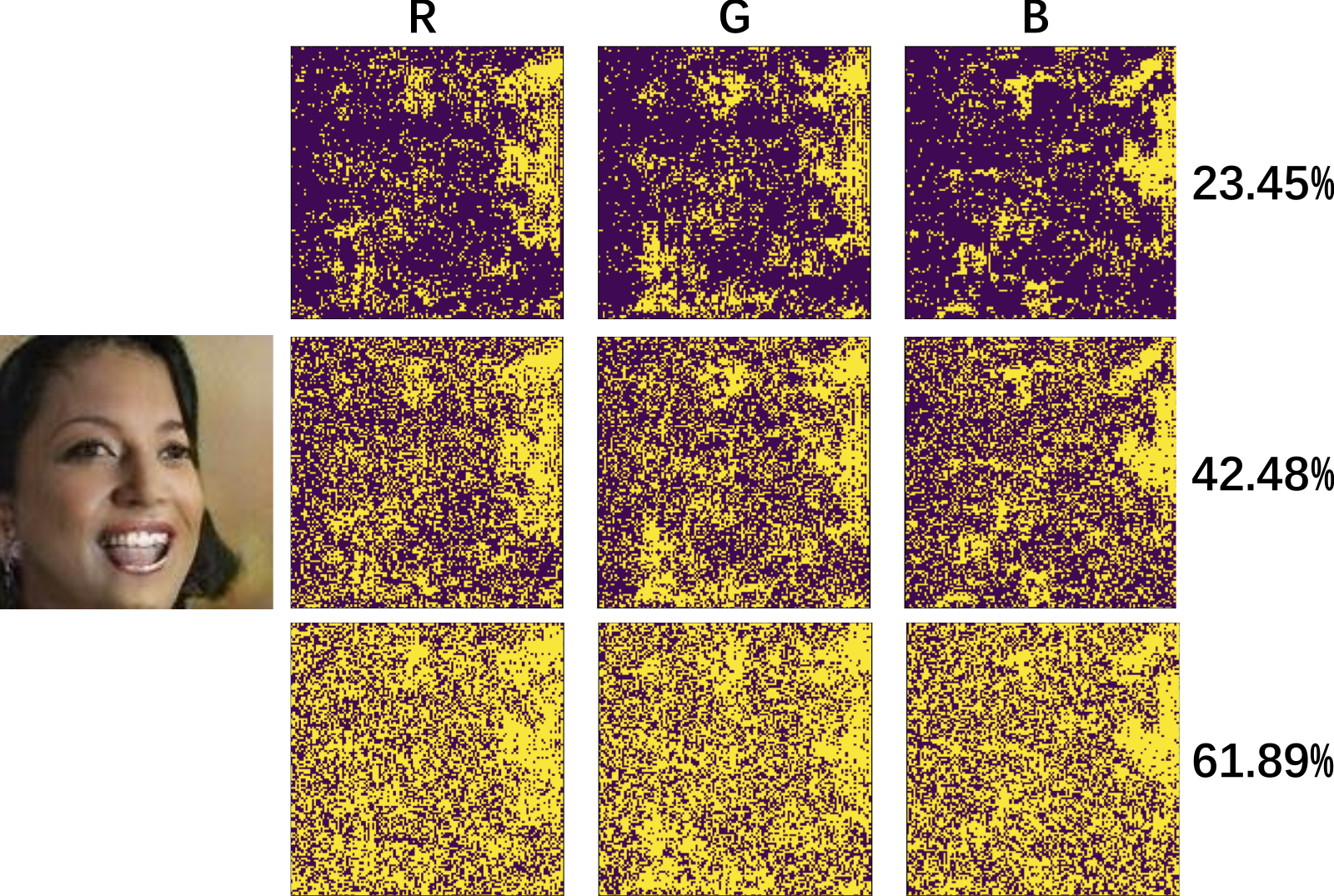}
\hspace{0.5in}
\includegraphics[scale=0.3]{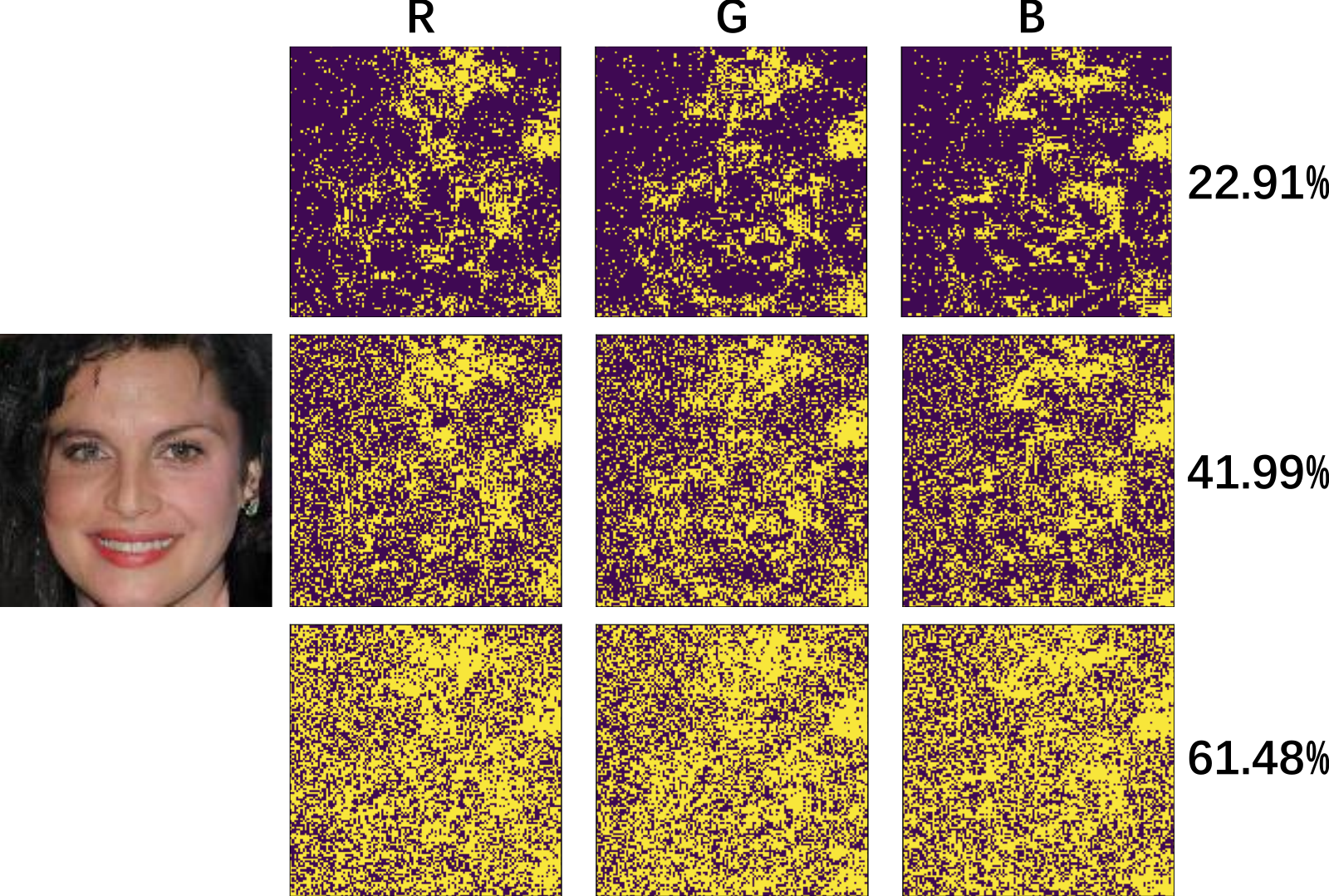}
\caption{The fake image in the first column of left subfigure and right subfigure are generated by ProGAN \cite{karras2017progressive} and SNGAN \cite{miyato2018spectral} with CelebA \cite{liu2018large}, respectively. The images on the right are the corresponding adversarial guided maps. The labels \textbf{R}, \textbf{G}, \textbf{B} in the first row represent red, green, and blue channels of the adversarial guided map. The adversarial guided maps under the labels are corresponding to one of the three channels of the fake image on the left (\eg, the red-channel adversarial guided map is used to confirm where to add noise to the red channel of the fake image). In each adversarial guided map, the pixel in it is either 0 or 1. The yellow pixel means 1 while the purple pixel means 0. The area of yellow pixels is where we will add noise to. The numbers in the last column mean the percentage of the yellow area in the whole three channels of adversarial guided maps in the same row.}
\label{fig:adversarial_guided_map_show}
\end{figure*}

\begin{figure}[]
	\centering 
	\includegraphics[width=0.95\linewidth]{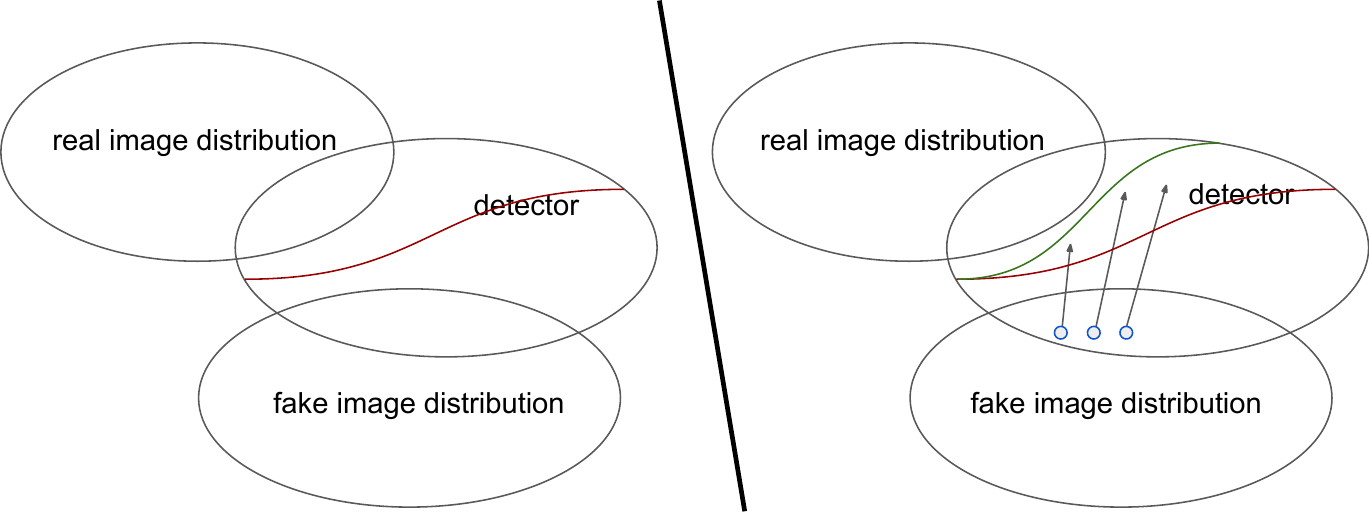}
	\caption{The left subfigure shows the ability of DeepFake detectors. The ability of the detectors has interacted with real image distribution and fake image distribution. Samples above the decision line (\ie, red line) of the detector are considered as real while samples below the decision line are considered as fake. The right subfigure shows that our method can shift the fake images (\ie, blue points) to be above the decision line (\ie, red line) of the detector. With our reconstructed images, the detectors can be retrained to have a more accurate decision line (\ie, green line).}
	\label{fig:detector_ability}
\end{figure}

\begin{table}
\centering
\caption{Similarity between fake image \& generated image of GANs in GANFingerprint \& DCTA.}
\resizebox{0.99\linewidth}{!}{
\begin{tabular}{c c c c c c}
\toprule
\multicolumn{2}{c}{} &  ProGAN & SNGAN & CramerGAN & MMDGAN\\ 
\toprule
\multirow{3}{*}{DN(rn)-gau} &  COSS   & 0.998 & 0.998 & 0.998 & 0.998   \\ 
&PSNR   & 30.26  & 30.21 & 30.18 & 30.26   \\ 
&SSIM  & 0.933 & 0.932 &  0.933 & 0.936 \\ \midrule
\multirow{3}{*}{DN(rn)-uni} &  COSS  & 0.998 & 0.998 & 0.998 & 0.998   \\
&PSNR   & 29.91 & 29.86 & 29.79 & 29.88  \\ 
&SSIM  & 0.928 & 0.926 & 0.927 & 0.930\\\midrule
\multirow{3}{*}{DN(an)-gau} &  COSS  & 0.998 & 0.998 & 0.998 & 0.998   \\
&PSNR   & 30.00 & 29.97 & 29.93 & 30.00  \\ 
&SSIM  & 0.929 & 0.928 & 0.929 & 0.932\\\midrule
\multirow{3}{*}{DN(an)-uni} &  COSS & 0.998 &0.998 & 0.998& 0.998  \\
&PSNR   & 29.73 & 29.68& 29.61&  29.70 \\ 
&SSIM  & 0.925& 0.924 &0.924 &0.927  \\\midrule
\multirow{3}{*}{StatAttack} &  COSS   & 0.973 &0.972 & 0.974& 0.973  \\
&PSNR   & 17.66 & 17.73& 17.71&  17.61 \\ 
&SSIM  & 0.840& 0.841 &0.841 &0.841  \\
\bottomrule
\end{tabular}
}
\label{Table:GANFingerprint_Similarity}
\end{table}

\begin{table*}[htb]
\centering
\caption{Similarity between fake image \& generated image of GANs in CNNDetector.}
\resizebox{1\linewidth}{!}{
\begin{tabular}{c c c c c c c c c c c c c c c c c c c c c c c}
\toprule

\multicolumn{2}{c}{\multirow{2}{*}{}}  & \cellcolor{gray!40} BigGan   & DeepFakes  & \cellcolor{gray!40} GauGAN  &IMLE  & \cellcolor{gray!40} SAN  & SITD  & \cellcolor{gray!40} StarGAN  & Whichfaceisreal  &  \multicolumn{6}{>{\columncolor{gray!40}}c}{CycleGAN} & \multicolumn{3}{c}{StyleGAN} &  \multicolumn{4}{>{\columncolor{gray!40}}c} {StyleGAN2}\\
                  \multicolumn{2}{c}{} & -  & person  & - & road & - & -  & person  & person  & horse & zebra & winter  & orange  & apple & summer & bedroom & car  & cat &horse & church & car & cat \\ 
                  
                  \toprule
\multirow{3}{*}{DN(rn)-gau} &  COSS          & 0.997  & 0.999  & 0.997 &  0.999 & 0.994 & 0.988  & 0.999  & 0.997 & 0.998 & 0.998 & 0.997  & 0.998  & 0.997 & 0.996 & 0.998  & 0.995  & 0.998 & 0.998  & 0.997  & 0.996 &  0.999    \\ 
&PSNR               & 28.95  & 32.73  & 28.93 & 32.02  & 27.19 & 28.09  &  32.10 & 28.82 & 30.11 & 28.94 &  28.44 & 29.67  & 29.56 & 28.22 & 29.88  & 26.85  & 30.61 & 29.51  & 27.43  & 27.61 & 31.87    \\ 
&SSIM               &  0.917 & 0.976  & 0.910 & 0.971  & 0.877 & 0.857  & 0.929  & 0.873 & 0.943 & 0.933 & 0.909  & 0.928  & 0.895 & 0.903 & 0.938  & 0.870  & 0.937 & 0.929  & 0.909  & 0.888 &  0.960   \\ \hline
\multirow{3}{*}{DN(rn)-uni}  &  COSS & 0.996  & 0.999 & 0.996  & 0.998 &  0.993 &  0.987 & 0.998 & 0.997 & 0.998 & 0.997  & 0.996  & 0.998 & 0.997 &  0.995 & 0.998  & 0.994 &  0.998 & 0.997  & 0.996 &  0.996  & 0.999 \\  
&PSNR   & 28.47  & 32.44  & 28.38 & 31.27  & 26.55 & 27.74  & 31.83  & 28.57 & 29.43 & 28.14 & 27.96  &  29.76 &29.25  & 27.73 & 29.26  & 26.30  & 30.16 & 28.93  & 26.89  & 27.04 & 31.31    \\  
&SSIM   & 0.907  & 0.975  & 0.900 & 0.965  & 0.863 & 0.853  & 0.926  & 0.867 & 0.932 & 0.920 & 0.899  & 0.913  & 0.884 &  0.892  & 0.930  & 0.856 & 0.931  &  0.917 & 0.894 & 0.875 & 0.955   \\ \hline
\multirow{3}{*}{StatAttack} & COSS & 0.967 & 0.971 & 0.964 & 0.965 & 0.958 & 0.939 & 0.973 & 0.970 & 0.966 & 0.965 & 0.966 & 0.974 & 0.972 & 0.965 & 0.972 & 0.964 & 0.968 & 0.964 & 0.972 & 0.961 & 0.966\tabularnewline
 & PSNR & 17.35 & 17.46 & 17.49 & 17.40 & 16.58 & 20.03 & 18.12 & 17.44 & 16.85 & 16.40 & 17.47 & 18.11 & 18.31 & 17.75 & 16.65 & 16.96 & 17.40 & 16.99 & 16.69 & 16.82 & 17.01\tabularnewline
 & SSIM & 0.812 & 0.896 & 0.814 & 0.860 & 0.738 & 0.796 & 0.875 & 0.801 & 0.800 & 0.773 & 0.787 & 0.851 & 0.843 & 0.784 & 0.833 & 0.769 & 0.838 & 0.793 & 0.782 & 0.777 & 0.850\tabularnewline
\toprule

\multicolumn{2}{c}{\multirow{2}{*}{}}  &  \multicolumn{20}{>{\columncolor{gray!40}}c}{ProGAN} & CRN\\ 
                  \multicolumn{2}{c}{} &  airplane & motorbike  & tvmonitor & horse & sofa  & car &  pottedplant & diningtable & sheep & bottle & person  & train  & dog  & cow & bicycle  & cat & bird  & boat & chair & bus & road \\
                  
                \toprule
                  
\multirow{3}{*}{DN(rn)-gau} &  COSS  &  0.998 & 0.996  & 0.997 &  0.997 & 0.998 & 0.997  & 0.996  & 0.997  & 0.997 & 0.998 &  0.997 & 0.997  & 0.998 & 0.997 &  0.996 & 0.998  & 0.997 & 0.997  &  0.998 & 0.997 & 0.999    \\ 
&PSNR               &  29.83 & 27.35  & 28.84 & 28.44  & 29.92 & 28.22  &  27.65 & 28.26 & 28.27 & 29.63 &  29.09 & 28.03  & 29.50  & 28.42 & 27.26  & 29.84  & 29.16 & 28.34  & 29.65  & 27.81 & 31.01    \\ 
&SSIM               &  0.947 & 0.922  & 0.917 & 0.918  & 0.942 & 0.916  &  0.920 & 0.921 & 0.908 & 0.930 &  0.920 & 0.916  & 0.927 & 0.909 & 0.918  & 0.930  & 0.933 &  0.922 & 0.938  & 0.920 & 0.966     \\ \hline
\multirow{3}{*}{DN(rn)-uni}&  COSS   & 0.998  & 0.996 & 0.997  & 0.997 &  0.998 & 0.996  & 0.996 & 0.996 & 0.996 & 0.997  & 0.997  & 0.996 & 0.997 &  0.997 & 0.995  & 0.998 &  0.997 &  0.997 & 0.997 &  0.996 & 0.998  \\  
&PSNR  & 29.11  & 26.63  & 28.06 & 27.80  & 29.15 & 27.47  & 26.87  & 27.43 & 27.69 & 28.92 & 28.50  & 27.31  & 28.98 & 27.83 & 26.49  &  29.28 & 28.58 & 27.57  & 28.73  & 27.00 &  30.27    \\  
&SSIM  & 0.940  & 0.911  & 0.909 &  0.906 & 0.933 & 0.905  & 0.905  & 0.908 & 0.895 & 0.921 &  0.912 & 0.902  & 0.919 & 0.897 & 0.904  &  0.923 & 0.923 & 0.910  & 0.928  & 0.907  &  0.958   \\ \hline
 \multirow{3}{*}{StatAttack} & COSS & 0.972 & 0.966 & 0.974 & 0.967 & 0.973 & 0.967 & 0.966 & 0.968 & 0.967 & 0.975 & 0.969 & 0.969 & 0.970 & 0.967 & 0.965 & 0.969 & 0.967 & 0.968 & 0.975 & 0.968 & 0.968\tabularnewline
 & PSNR & 16.94 & 16.76 & 17.58 & 16.96 & 16.93 & 17.18 & 16.69 & 16.82 & 17.12 & 17.09 & 17.63 & 17.21 & 17.21 & 17.05 & 16.94 & 17.21 & 17.07 & 16.60 & 17.10 & 16.95 & 17.11\tabularnewline
 & SSIM & 0.836 & 0.786 & 0.831 & 0.795 & 0.830 & 0.803 & 0.772 & 0.792 & 0.775 & 0.838 & 0.823 & 0.789 & 0.824 & 0.788 & 0.777 & 0.829 & 0.815 & 0.791 & 0.837 & 0.791 & 0.851\tabularnewline
\bottomrule
\end{tabular}}
\label{Table:CNNDetector_Similarity}
\end{table*}

\subsection{Discussion}\label{sec:discussion}

\subsubsection{Similarity b/w fake and reconstructed image}
\begin{figure}
	\centering 
	\includegraphics[width=0.49\textwidth]{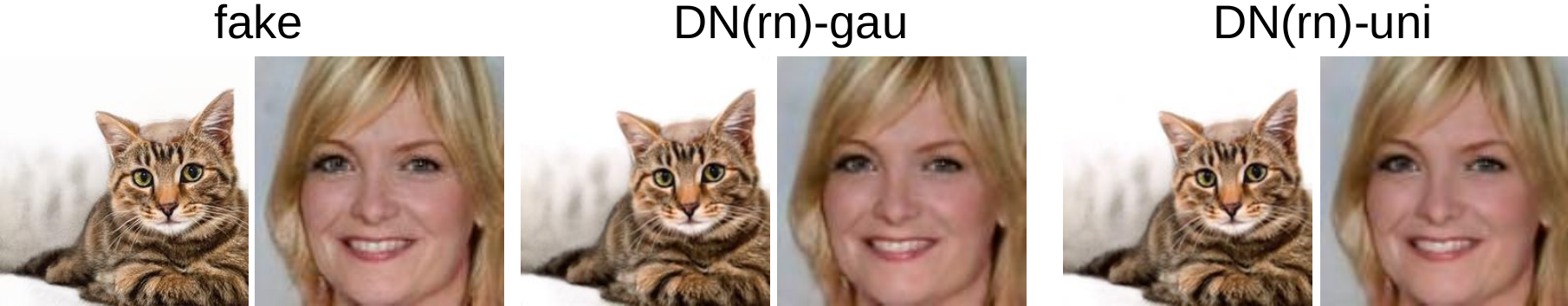}
	\caption{Reconstructed images of different categories. The image pairs in turn are fake images, DN(rn)-gau reconstructed image and DN(rn)-uni reconstructed image of cat and human. The fake image of the cat is produced by StyleGAN \cite{karras2019style}. The fake image of humans is produced by ProGAN \cite{karras2017progressive}.}
	\label{fig:reconstruction_show}
\end{figure}

Table \ref{Table:GANFingerprint_Similarity} summarizes the similarity between fake images and reconstructed images of GANs in GANFingerprint and DCTA. For each of DN(rn)-gau, DN(rn)-uni and DN(an)-uni, we use three metrics (COSS, PSNR, SSIM) to demonstrate the similarity. The values of COSS and SSIM of all the GANs are close to 1, indicating a high similarity between fake images and reconstructed images. The values of PSNR are also very high.
For CNNDetector, the similarity of DN(rn)-gau/DN(rn)-uni reconstructed images and fake images are summarized in Table \ref{Table:CNNDetector_Similarity}, where 13 GANs have different subclasses. 
Among these, five GANs contain only one category. The images in DeepFakes, StarGAN, and Whichfaceisreal are all persons while the images in IMLE and crn are both roads. 
The other eight GANs have multiple categories. The images of CycleGAN, StyleGAN, StyleGAN2, and ProGAN are sorted into categories, they use different folders to store different categories of images. The other GANs (BigGAN, GauGAN, SAN, SITD) combine images of different categories into one folder. We use `-' to represent this category. We can see that the reconstructed images in CNNDetector are also very similar to their fake image counterparts.
Fig.~\ref{fig:reconstruction_show} shows the reconstructed image of DeepNotch on a cat and a human face, with very high reconstruction quality.

\subsubsection{Discussion on deliberate noise}
Fig.~\ref{fig:region_compare} shows the performance between deliberate noise and random noise of the same noise setting (\ie, uniform noise with 20 upper bound and -20 lower bound). The horizontal axis of all four subfigures represents the percentage of the region ratio of the image, on which the noise is added. The vertical axis is the possibility of reconstructed images being classified as {CelebA} types by method {GANFingerprint}. The blue and orange lines represent the performance of deliberate noise and random noise, respectively. We can observe that in all the four GANs and among the different percentages of regions, the accuracy of deliberate noise is stably better than that of random noise. This indicates that our localization method is stable and effective. 

\subsubsection{Discussion on adversarial guided map}
As shown in the left subfigure  of Fig.~\ref{fig:adversarial_guided_map_show}, the fake image in the first column is generated by ProGAN \cite{karras2017progressive} with CelebA \cite{liu2018large}.
The images on the right are the adversarial guided maps. The labels \textbf{R}, \textbf{G}, \textbf{B} in the first row represent the red, green, and blue channels of the adversarial guided map. The adversarial guided maps under the labels are corresponding to one of the three channels of the fake image on the left (\eg, the red-channel adversarial guided map is used to confirm where to add noise to the red channel of the fake image). In each adversarial guided map, the pixel in it is either 0 or 1. The yellow pixel means 1 while the purple pixel means 0. The area of yellow pixels is where we will add noise to. The numbers in the last column mean the percentage of the yellow area in the whole three channels of adversarial guided maps in the same row. As shown in the right subfigure of  Fig.~\ref{fig:adversarial_guided_map_show}, the fake image in the first column is generated by SNGAN \cite{miyato2018spectral} with CelebA \cite{liu2018large}. We can find that the adversarial guided maps are semantic-aware.

\begin{figure}[]
    \centering 
    \includegraphics[width=\linewidth]{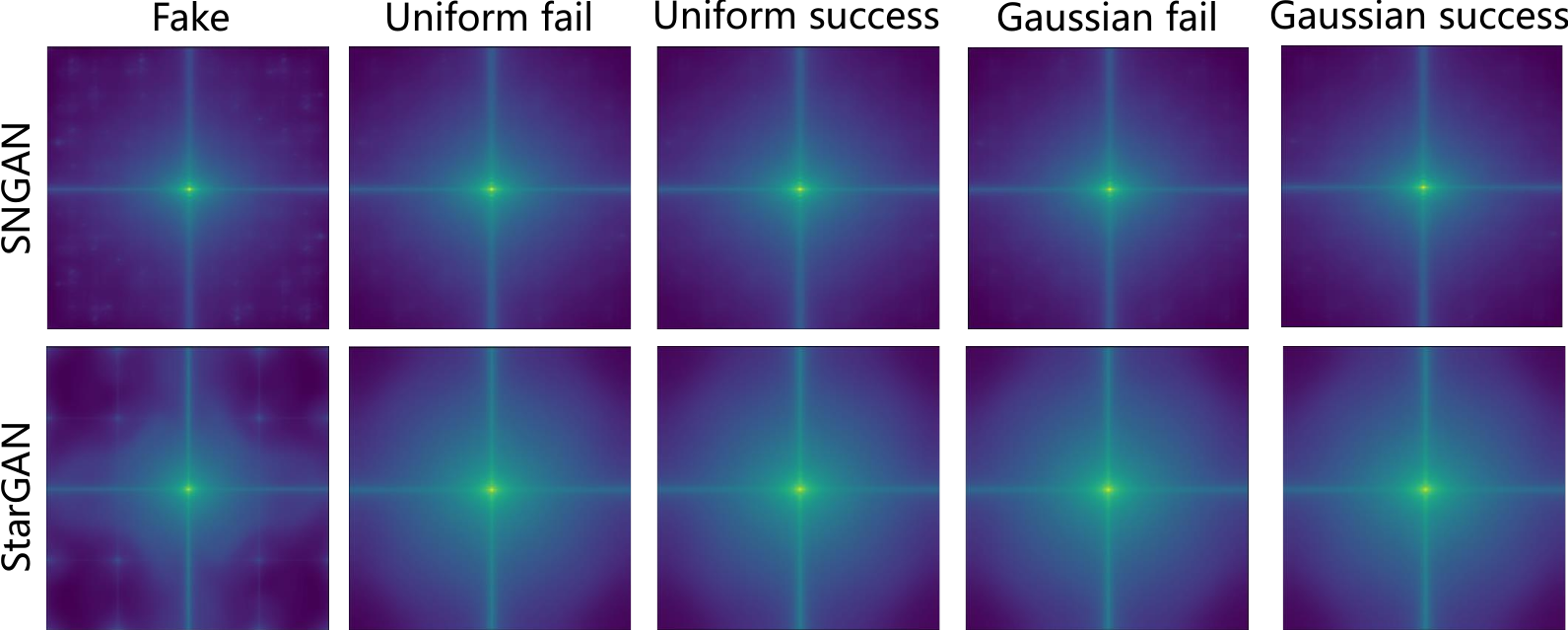}
    \caption{Spectrums of failure samples and success samples when evading detectors. There is no obvious difference between failure samples and success samples. This means other evidence may be captured by detectors.}
    \label{fig:failure_success_samples_spectrum}
\end{figure}

\subsubsection{Discussion on failure samples}
We investigate the images that can still detected as DeepFakes despite undergoing the DeepNotch and call them ``failure samples''. In contrast, the images that are detected as Real undergoing the DeepNotch are called ``success samples''. We find that the spectrum of fail samples and success samples are similar and both almost show no artifacts. Please see the following Figure \ref{fig:failure_success_samples_spectrum}, we select SNGAN and StarGAN as examples since the spectrums of their fake images show obvious artifacts (\ie, blobs), which is easy for us to observe. The SNGAN is detected by GANFingerprint and the StarGAN is detected by CNNDetector. Let's take the first row as an example. The fake SNGAN images exhibit obvious artifacts in the spectrum. It's important to note that the spectrum is computed by averaging the spectra of fake SNGAN images. When we apply DeepNotch and uniform noise to filter the SNGAN images, we gather failure samples and success samples. Subsequently, we calculate the average spectrum for both sets, denoting them as "Uniform fail" and "Uniform success" in Figure \ref{fig:failure_success_samples_spectrum}. Remarkably, we find that there are hardly any artifacts in either the failure or success samples. Similarly, when we filter the SNGAN images using DeepNotch and Gaussian noise, we observe almost no artifacts in both sets of samples. We can also find that in the second row (StarGAN), there are almost no artifacts in the spectrums of fail samples and success samples. This finding implies that, in addition to the artifacts discussed in our paper, there may exist other indicators capable of distinguishing fake from real images. However, since DeepNotch effectively reduces artifacts and allows evasion of existing DeepFake detectors like GANFingerprint, DCTA, and CNNDetector, we believe that these alternative indicators have limited significance in current DeepFake detectors. It would be intriguing to identify these additional indicators, and we intend to investigate them in our future research.

\subsubsection{Comparison with adversarial attack methods}
\begin{figure}[t]
    \centering 
    \includegraphics[width=0.8\linewidth]{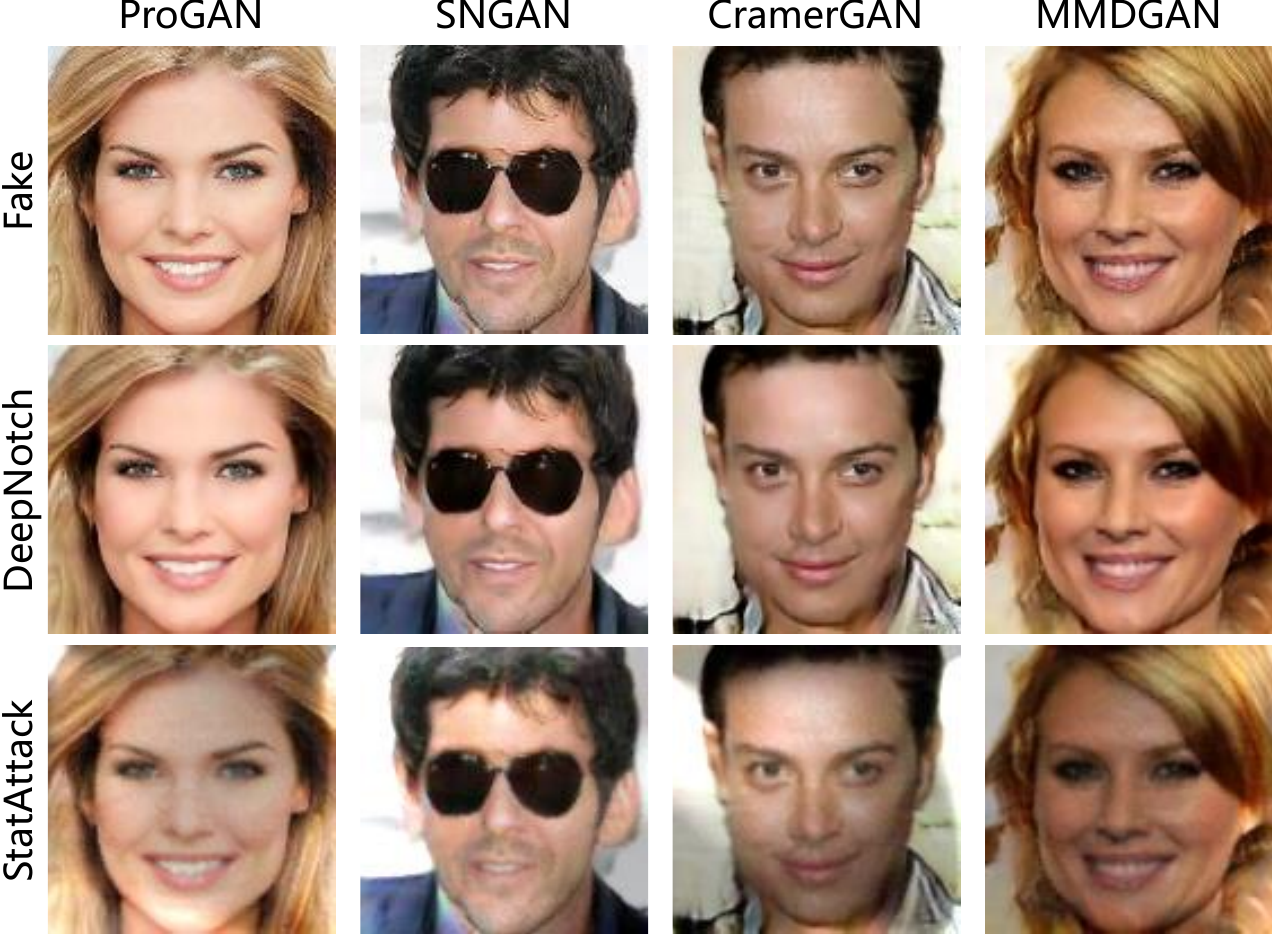}
    \caption{Visualization of images generated by DeepNotch and StatAttack. We can find that the images generated by StatAttack have obvious corruptions which may raise suspicion.}
    \label{fig:DeepNotch_StatAttack_Vis}
\end{figure}

We further investigate the SOTA detector evasion methods and find that currently the most popular methods are based on adversarial attacks. Although the adversarial attack is another distinct method compared with image reconstruction methods, in order to show the comparison comprehensively, we add extra experiments. To our best knowledge, the SOTA published methods are TR-Net \cite{liu2023making} and StatAttack \cite{hou2023evading}. TR-Net does not open-source the code. Compared with that, StatAttack open-source codes on GitHub\footnote{https://github.com/tobuta/evadingfakedetector}. Thus we conduct experiments of StatAttack with the same attack setting in \cite{hou2023evading} and the same experiment setting in our paper on three different DeepFake detectors. Since DeepNotch is a target detector-independent evasion method, we apply the StatAttack method in the black-box setting. As shown in Table~\ref{Table:GANFingerprint_Acc}, when evading GANfingerprint, we can find that DeepNotch shows better performance than StatAttack. As shown in Table~\ref{Table:DCTA_Acc}, when evading DCTA, we can find that StatAttack is a bit better than DeepNotch. As shown in Table~\ref{Table:CNNDetector_Acc}, when evading CNNDetector, we can find that StatAttack is better than DeepNotch on most of the DeepFakes. Furthermore, we compare the image quality of images generated by DeepNotch and StatAttack. As shown in Table~\ref{Table:GANFingerprint_Similarity} and Table~\ref{Table:CNNDetector_Similarity}, we can find that the images generated by DeepNotch are more similar to the corresponding original images than those generated by StatAttack. As shown in Fig.~\ref{fig:DeepNotch_StatAttack_Vis}, we take the DeepFake dataset of GANFingerprint and DCTA as an example. We can find that the images generated by StatAttack have obvious corruption, which makes the images seem uncommon. To sum up, compared with SOTA DeepFake evasion methods (\ie, FakePolisher) that are based on image reconstruction, DeepNotch shows better evasion performance. Compared with the SOTA DeepFake evasion method (\ie, StatAttack) of different types we can find that DeepNotch shows better image quality while a bit poor evasion performance.

\subsubsection{Limitation}
As shown in Table \ref{Table:GANFingerprint_Retrain_Acc}, people could retrain the DeepFake detection model to defend our method, which is the limitation of our method. However, this means that our method can help to improve the DeepFake detector to a new version. Furthermore, improved detectors will encourage researchers to find more effective attack methods, which brings continuous progress in the DeepFake field.

\subsubsection{Target of our method}
In our point of view, the DeepFake detectors are similar to the discriminators of GAN. As shown in the left of Fig.~\ref{fig:detector_ability}, their classification ability has an intersection with part of the real distribution and fake distribution. If we use a red line to divide the ability of the detector into two parts, then we can find that the images above the red line will be considered as real, although they may not be actually real. Similarly, the images below the red line will be considered fake, although they may not be actually fake. From this point of view, we can find that the detectors still have room for improvement. The target of our method is to expose this defect with hard examples and help to improve the detectors with these hard examples. As shown in the right of Fig.~\ref{fig:detector_ability}, our method can shift the fake images (\ie, blue points) to be above the red line, which means the reconstructed images are considered as real by detectors. As shown in Table \ref{Table:GANFingerprint_Retrain_Acc}, with our reconstructed images (\ie, hard examples), the detector can be retrained to successfully classify reconstructed images and real images. This means our reconstructed images help the detectors to find a better decision line (\ie, green line) to classify real and non-real images. This shows the function of our method to promote the improvement of the DeepFake field.

\section{Conclusions}\label{sec:concl}
In this paper, we propose the \textbf{DeepNotch}, a pipeline that performs implicit spatial-domain notch filtering by taking a hybrid approach of deep image filtering and noise addition for improving the fidelity of GAN-based fake images.
Our method effectively reduces the artifact patterns introduced by existing fake image generation methods, in both spatial and frequency domains.
By reducing fake artifacts, our further reconstructed image retains photo-realistic and high-fidelity, which can bypass state-of-the-art DeepFake detection methods. Our large-scale evaluation demonstrates that more general DeepFake detectors beyond leveraging fake artifacts should be further investigated. In future work, we aim to learn from other video-related work \cite{chen2022saliency,liu2022temporal,shu2021large,deng2020spatio,chen2021learning,afonso2018video} for constructing DeepFake evasion with more temporal information.

\bibliographystyle{IEEEtran}
\bibliography{ref}

\begin{IEEEbiography}[{\includegraphics[width=1in,height=1.25in,clip]{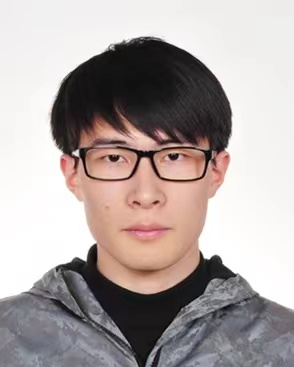}}]
{Yihao Huang} received the B.S. degree and Ph.D. degree in Software Engineering Institute, East China Normal University, China in 2017 and 2022. He is currently a research fellow with Nanyang Technological University. His research interests include computer vision and AI security. 
\end{IEEEbiography}

\vspace{-3em}
\begin{IEEEbiography}[{\includegraphics[width=1in,height=1.25in,clip]{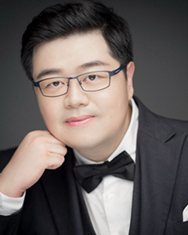}}]{Felix Juefei-Xu} (Member, IEEE) received the Ph.D. degree in Electrical and Computer Engineering from Carnegie Mellon University (CMU), Pittsburgh, PA, USA. Prior to that, he received the M.S. degree in Electrical and Computer Engineering and the M.S. degree in Machine Learning from CMU, and the B.S. degree in Electronic Engineering from Shanghai Jiao Tong University (SJTU), Shanghai, China. Currently, he is a Research Scientist with GenAI at Meta, based in New York City, where he works on robust perception and efficient learning problems in the domain of generative AI. He is also affiliated with New York University as an Adjunct Professor. Previously, he was a Research Scientist with Alibaba Group, based in Sunnyvale, CA. He was the recipient of multiple best/distinguished paper awards, including IJCB 2011, BTAS 2015 and 2016, ASE 2018, and ACCV 2018.
\end{IEEEbiography}

\vspace{-3em}
\begin{IEEEbiography}[{\includegraphics[width=1in,height=1.25in,clip,keepaspectratio]{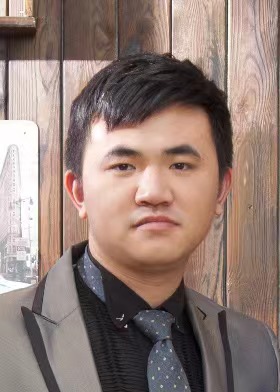}}]{Qing Guo} received his Ph.D. degree in computer application technology from the School of Computer Science and Technology, Tianjin University, China. He is currently a senior research scientist and principal investigator (PI) at the Center for Frontier AI Research (CFAR), A*STAR in Singapore. He is also an adjunct assistant professor at the National University of Singapore (NUS), and senior PC member of AAAI. Before that, he was a Wallenberg-NTU Presidential Postdoctoral Fellow with the Nanyang Technological University, Singapore. His research interests include computer vision, AI security, and image processing. He is a member of IEEE. 
\end{IEEEbiography}

\vspace{-3em}
\begin{IEEEbiography}[{\includegraphics[width=1in,height=1.25in,clip,keepaspectratio]{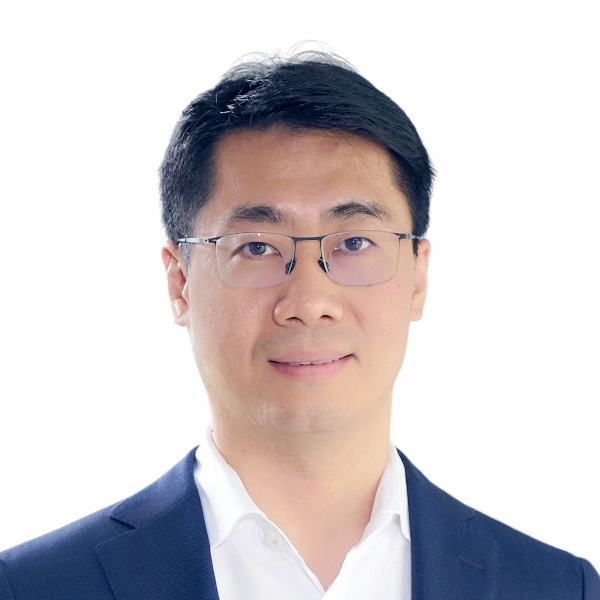}}]{Yang Liu}
graduated in 2005 with a Bachelor of
Computing (Honours) in the National University
of Singapore (NUS). In 2010, he obtained his
Ph.D. and started his post-doctoral work in NUS
and MIT. In 2012, he joined Nanyang Technological University (NTU), and currently is a full
professor and Director of the cybersecurity lab in
NTU.
Dr. Liu specializes in software engineering, cybersecurity and artificial intelligence. His research has bridged the gap between the theory and practical usage of program analysis, data analysis and AI to evaluate the design and implementation of software for high assurance and security. By now, he has more than 400 publications in top tier conferences and journals. He has received a number of prestigious awards including MSRA Fellowship, TRF Fellowship, Nanyang Assistant Professor, Tan Chin Tuan Fellowship, Nanyang Research Award 2019, ACM Distinguished Speaker, NRF Investigatorship, and 15 best paper awards and one most influence system award in top software engineering conferences like ASE, FSE and ICSE.
\end{IEEEbiography}

\vspace{-3em}
\begin{IEEEbiography}[{\includegraphics[width=1in,height=1.25in,clip,keepaspectratio]{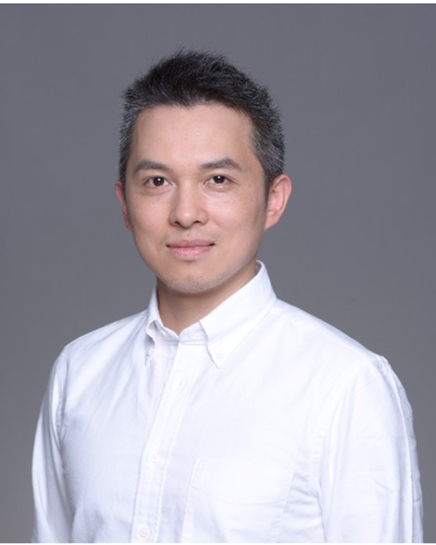}}]{Geguang Pu} is a Professor in Software Engineering Institute, East China Normal University. His research interests include program testing and reliable AI system. He served as PC member for more than 20 international conference committees. He has published over 100 publications on the topics of software engineering and system verification (including ICSE, FSE, ASE, CAV, etc). He completed his Ph.D. in Mathematics at Peking University in 2005 , and received a B.S. in Mathematics from Wuhan University in 2000. 
\end{IEEEbiography}

\end{document}